\def\confName{xxx}
\def\confYear{xxx}

\def\paperTitle{DiTCtrl: Exploring Attention Control in Multi-Modal Diffusion Transformer \\ for Tuning-Free Multi-Prompt Longer Video Generation}

\def\authorBlock{
    Minghong Cai$^{1\dagger}$  \qquad
    Xiaodong Cun$^2$ \qquad
    Xiaoyu Li$^3$\Envelope \qquad
    Wenze Liu$^1$ \qquad
    Zhaoyang Zhang$^3$ \\
    \vspace{0.2cm}
    Yong Zhang$^4$ \qquad
    Ying Shan$^{3}$ \qquad
    Xiangyu Yue$^{1,5}$\Envelope \\
    
    $^1$~MMLab, CUHK \qquad
    $^2$~GVC Lab, Great Bay University \vspace{0.1cm}\\
    $^3$~ARC Lab, Tencent PCG \qquad
    $^4$~Tencent AI Lab \qquad
    $^5$~SHIAE, CUHK \vspace{0.2cm} 
}

\newif\ifreview 
\newif\ifarxiv \newcommand{\arxiv}{\arxivtrue}
\newif\ifcamera 
\newif\ifrebuttal 

\arxiv % \review OR \arxiv OR \cameraready

\pdfoutput=1
\documentclass[10pt,twocolumn,letterpaper]{article}
\ifreview \usepackage[review]{cvpr} \fi
\ifarxiv \usepackage[pagenumbers]{cvpr} \fi
\ifrebuttal \usepackage[rebuttal]{cvpr} \fi
\ifcamera \usepackage{cvpr} \fi
\def\paperID{11688} % *** Enter the Paper ID here
\def\confName{CVPR}
\def\confYear{2025}

\usepackage{graphicx}
\usepackage{amsmath}
\usepackage{amssymb}
\usepackage{booktabs}

\usepackage{times}
\usepackage{microtype}
\usepackage{epsfig}
\usepackage{caption}
\usepackage{float}
\usepackage{placeins}
\usepackage{color, colortbl}
\usepackage{stfloats}
\usepackage{enumitem}
\usepackage{tabularx}
\usepackage{xstring}
\usepackage{multirow}
\usepackage{xspace}
\usepackage{url}
\usepackage{subcaption}
\usepackage[hang,flushmargin]{footmisc}
\usepackage{bbding}

\ifcamera \usepackage[accsupp]{axessibility} \fi

% ...

\definecolor{alizarin}{rgb}{0.82, 0.1, 0.26}

\ifarxiv  \fi

\usepackage{xargs}
\usepackage[colorinlistoftodos,prependcaption,textsize=normalsize]{todonotes}
\newcommandx{\xiangyu}[2][1=]{\todo[linecolor=red,backgroundcolor=red!25,bordercolor=red,#1]{#2}}

\newcommand{\R}[1]{{%
    \textbf{%
        \ifstrequal{#1}{1}{\textcolor{red}{R#1}}{%
        \ifstrequal{#1}{2}{\textcolor{blue}{R#1}}{%
        \ifstrequal{#1}{3}{\textcolor{magenta}{R#1}}{%
        \ifstrequal{#1}{4}{\textcolor{teal}{R#1}}{%
                           \textcolor{cyan}{R#1}%
        }}}}%
    }%
}}

\usepackage[ruled,vlined]{algorithm2e}
\SetAlgoLined
\SetKwInput{KwIn}{Input}
\SetKwInput{KwOut}{Output}
\SetKwComment{tcp}{/* }{ */}
\usepackage{xr-hyper}

\makeatletter
\newcommand*{\addFileDependency}[1]{
  \typeout{(#1)}
  \@addtofilelist{#1}
  \IfFileExists{#1}{}{\typeout{No file #1.}}
}

\makeatother

\usepackage[pagebackref,breaklinks,colorlinks]{hyperref}
\usepackage[capitalize]{cleveref}
\crefname{section}{Sec.}{Secs.}
\crefname{table}{Table}{Tables}
\crefname{figure}{Fig.}{Figs.}

\hypersetup{
    colorlinks = true,
    linkbordercolor = {white},
    citecolor=blue,
    urlcolor=alizarin
}

\begin{document}
\marginparwidth=\dimexpr \marginparwidth + 1.5cm\relax
%% TITLE
\title{\paperTitle}
\author{\authorBlock}

% \maketitle
%%

% \maketitle
\twocolumn[{
\maketitle
\begin{center}
    \captionsetup{type=figure}
    \vspace{-2em}
    \includegraphics[width=1.\textwidth]{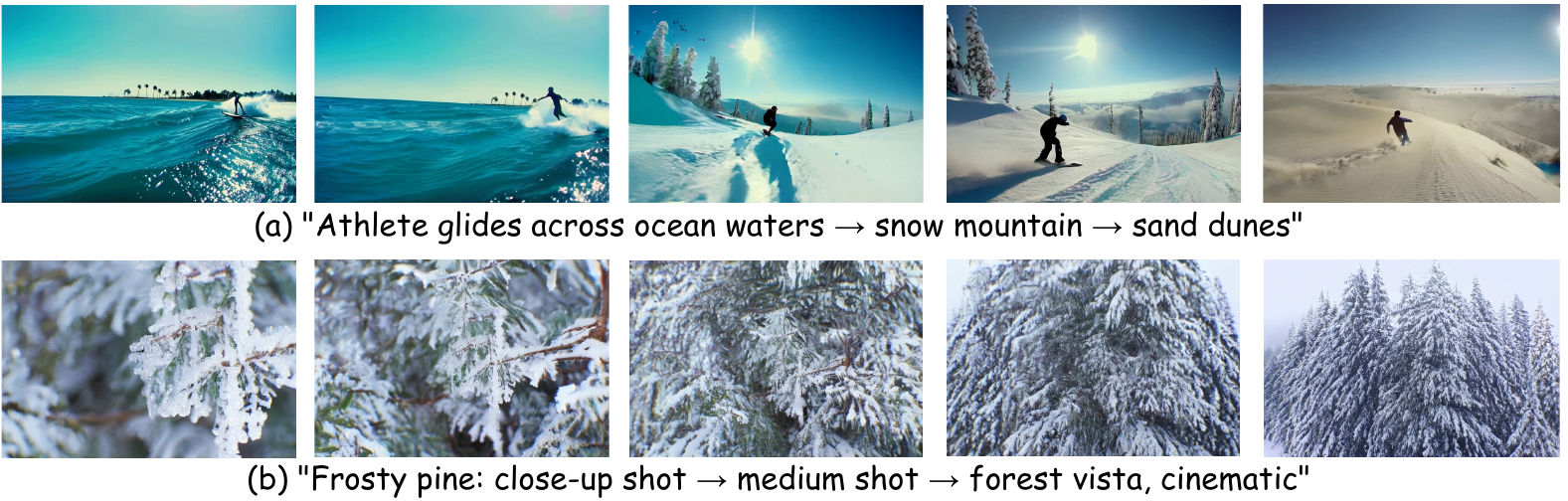}
    \vspace{-2em}
    \captionof{figure}{
    Our method \textbf{DiTCtrl} takes multiple text prompts as input and demonstrates superior capability in generating longer videos with multiple events, long-range coherence and smooth transitions as output. 
    }
    \label{fig:teaser}
\end{center}
}]

\let\thefootnote\relax\footnotetext{$\dagger$ Work done during an internship at Tencent ARC Lab.\\
\Envelope~Corresponding Author
}
\begin{abstract}

{Sora-like video generation models have achieved remarkable progress with a Multi-Modal Diffusion Transformer~(MM-DiT) architecture.}
However, the current video generation models predominantly focus on single-prompt, struggling to generate coherent scenes with multiple sequential prompts that better reflect real-world dynamic scenarios. 
While some pioneering works have explored multi-prompt video generation, they face significant challenges including strict training data requirements, weak prompt following, and unnatural transitions. To address these problems, we propose DiTCtrl, a training-free multi-prompt video generation method under MM-DiT architectures for the first time. Our key idea is to take the multi-prompt video generation task as temporal video editing with smooth transitions. To achieve this goal, we first analyze MM-DiT's attention mechanism, finding that the 3D full attention behaves similarly to that of the cross/self-attention blocks in the UNet-like diffusion models, enabling mask-guided precise semantic control across different prompts with attention sharing for multi-prompt video generation. 
Based on our careful design, the video generated by DiTCtrl achieves smooth transitions and consistent object motion given multiple sequential prompts without additional training. 
Besides, we also present MPVBench, a new benchmark specially designed for multi-prompt video generation to evaluate the performance of multi-prompt generation. Extensive experiments demonstrate that our method achieves state-of-the-art performance without additional training. 
Code is available at \href{https://github.com/TencentARC/DiTCtrl}{{https://github.com/TencentARC/DiTCtrl}}.

\end{abstract}

\section{Introduction}
\label{sec:intro}

Text-to-video~(T2V) generation has made remarkable progress in the AIGC era~\cite{chen2023videocrafter1, zhou2022magicvideo, wang2023modelscope}, and breakthroughs such as Sora~\cite{sora} have demonstrated impressive capabilities in generating longer videos through DiT~\cite{dit} architecture and large-scale per-taining. 
% Additionally, state-of-the-art open-source models (\eg, CogVideoX~\cite{yang2024cogvideox}, Mochi1~\cite{mochi1}) have pushed the boundaries of video generation.
% Naturally, these approaches operate under a single-prompt paradigm. 
However, feeding \textit{sequential} prompts into current state-of-the-art text-to-video generation models~(\eg, Kling~\cite{kuaishou}, Gen3~\cite{runway}, CogVideoX~\cite{yang2024cogvideox}) directly will produce isolated video sequences without natural transitions, as shown in Fig.~\ref{fig: fig_qualitative} (First row). On the other hand, when multiple prompts are consolidated into a single-prompt describing long-term temporal changes, the generated results fail to capture semantic transitions effectively, as demonstrated in Fig.~\ref{fig: fig_qualitative} (Second row). This limitation stems from their fundamental design and single-prompt training paradigm, making them inadequate for depicting real-world scenarios' dynamic, multi-event nature. 

Although pioneering works~\cite{bansal2024talc,oh2023mtvg,villegas2022phenaki} have begun exploring multi-prompt video generation, they face significant challenges. \eg, training such extended video generation models~\cite{villegas2022phenaki, oh2023mtvg} from scratch would require unprecedented computational resources and datasets that are practically unfeasible when the model size increases. Current zero-shot longer video generation methods~\cite{qiu2023freenoise,wang2023gen,kim2024fifo} still mainly focus on the single prompt situation with longer length.
% including intensive training data requirements and limited object motion. 
Moreover, all previous works~\cite{bansal2024talc,qiu2023freenoise,villegas2022phenaki,oh2023mtvg,kim2024fifo} are specifically designed under UNet architecture which restricts the abilities of more complex motions and increase the difficulties in multi-prompt generation. 
However, since Sora 's~\cite{sora} groundbreaking demonstration of two-minute video generation, highlighting the scalability potential of DiT architectures~\cite{dit}. Subsequent explorations have led to influential developments, notably in image generation models~(Stable Diffusion 3~\cite{sd3}, FLUX.1~\cite{flux}) and video generations~(CogVideoX~\cite{yang2024cogvideox}, Mochi1~\cite{mochi1}). They~\cite{sd3,flux,yang2024cogvideox,mochi1} all adopt a specific kind of DiT architecture, \ie, Multi-Modal Diffusion Transformer~(MM-DiT~\cite{sd3}) as the basic unit. This architecture effectively maps text and images~(or video) into a unified sequence for attention computation, enabling deeper model scale abilities and achieving superior performances.

\if
The emergence of DiT-based architectures has gained significant momentum, particularly since Sora 's~\cite{sora} groundbreaking demonstration of two-minute video generation, highlighting the scalability potential of DiT architectures~\cite{dit}. Subsequent explorations have led to influential developments, notably in image generation models like Stable Diffusion 3~\cite{sd3} and FLUX.1~\cite{flux}, which adopt the MM-DiT architecture. This architecture effectively maps text and images into a unified sequence for attention computation, achieving superior text-image alignment. Similarly, the current most successful open-source video generation models, CogVideoX~\cite{yang2024cogvideox} and Mochi1~\cite{mochi1}, have adapted comparable MM-DiT architectures, just replacing image-text combinations with video-text pairs. Despite this trend, current approaches to both single-prompt and multi-prompt long video generation still predominantly rely on UNet architectures, with limited exploration of DiT architectures' potential, particularly regarding MM-DiT applications. Our DiTCtrl method pioneers a more granular analysis of the MM-DiT full attention architecture, decomposing it into UNet-like partial cross-attention and partial self-attention components. Through semantic mapping in partial cross-attention, we achieve enhanced control over partial self-attention, significantly improving the quality of our multi-prompt video generation.

Our approach represents the first tuning-free implementation of multi-prompt video generation based on the DiT architecture. We introduce an overlap blending strategy to connect different semantic segments seamlessly. Inspired by \cite{hertz2022prompt,cao2023masactrl}, we conceptualize the transitions between semantic segments as an editing task, incorporating KV-sharing Attention methods from image editing to maintain semantic consistency of key objects across video segments. Furthermore, we employ mask-guided generation, extracting semantic map information from MM-DiT's partial cross-attention to enhance foreground-background consistency. This comprehensive approach enables the generation of long-form videos with smooth prompt transitions, characterized by consistent foreground object motion, character continuity, and fluid action sequences, all without requiring additional training or computational resources. To systematically evaluate our method and facilitate future research in multi-prompt video generation, we also introduce MPVBench, a new benchmark with diverse transition types and specialized metrics for assessing multi-prompt transitions. Extensive experiments on this benchmark demonstrate that our method achieves state-of-the-art performance while maintaining computational efficiency.
\fi

Thus, to keep the abilities of the pre-trained single prompt T2V model and take advantage of the performance of the diffusion transformer, we propose DiTCtrl, a \textit{training-free} multi-prompt video generation method under the pre-trained MM-DiT video generation model.
{Our key observation is that the multi-prompt video generation can be considered a two-step problem: 
\textit{1) Video editing over time: The new video is generated through the previous video with a new prompt. }
\textit{2) Video transition over time: Two generated videos need to keep a smooth transition between clips.}
Thus, to perform consistent video editing, inspired by the UNet-based image editing techniques~\cite{cao2023masactrl,hertz2022prompt},
% However, there is no explicit representation of the cross- and self-attentions in the MM-DiT block as a control handle. 
we explore the characteristic of the attention modules in the MM-DiT block for the first time, finding that the 3D full attention has similar behaviors to that of the cross-/self-attention blocks in the UNet-like diffusion models~\cite{chen2023videocrafter1, wang2023modelscope}. We thus apply a KV-sharing method between the video clips of different prompts to maintain the semantic consistency of the key objects~\cite{cao2023masactrl} with the 3D attention control. 
Besides, we utilize a latent blending strategy for transitions between clips to connect the video clip seamlessly. Finally, to systematically evaluate our method and facilitate future research in multi-prompt video generation, we also introduce MPVBench, a new benchmark with diverse transition types and specialized metrics for assessing multi-prompt transitions. Extensive experiments on this benchmark demonstrate that our method achieves state-of-the-art performance while maintaining computational efficiency.
}

The contributions of this paper can be summarized as:

\begin{itemize}
    \item We propose DiTCtrl, the first tuning-free approach based on MM-DiT architecture for coherent multi-prompt video generation. Our method incorporates a novel KV-sharing mechanism and latent blending strategy, enabling seamless transitions between different prompts without additional training.
    
    \item We pioneer the analysis of MM-DiT's attention mechanism, finding that 3D full attention has similar behaviors to that of the cross/self-attention blocks in the UNet-like diffusion models, enabling mask-guided precise semantic control across different prompts for enhanced generation consistency.
    \item We introduce MPVBench, a new benchmark specially designed for multi-prompt video generation, featuring diverse transition types and specialized metrics for multi-prompt video evaluation.
    \item Extensive experiments demonstrate that our method achieves state-of-the-art performance on multi-prompt video generation while maintaining computational efficiency.
\end{itemize}

\section{Related Work}
\label{sec:related}

\paragraph{Video Diffusion Model.}
Diffusion models have achieved significant success in the field of text-to-image generation~\cite{nichol2021glide, ramesh2022hierarchical, saharia2022photorealistic, rombach2022high}, and these advancements have also propelled progress in video generation from text or images~\cite{guo2023animatediff, ho2022imagen, singer2022make, chen2023videocrafter1, chen2024videocrafter2, blattmann2023stable, blattmann2023align, an2023latent, girdhar2023emu}. Among these methods, AnimateDiff~\cite{guo2023animatediff} attempts to turn existing text-to-image diffusion models with a motion module. Other models such as Imagen Video~\cite{ho2022imagen} and Make-a-Video~\cite{singer2022make} train a cascade model of spatial and temporal layers directly in pixel space. To improve efficiency, many other works~\cite{chen2023videocrafter1, chen2024videocrafter2, blattmann2023stable, blattmann2023align, an2023latent, girdhar2023emu} generate the videos in latent space, leveraging an auto-encoder to compress the video into a compact latent. Notably, most of these text-to-video models utilize a U-Net architecture. Subsequently, the introduction of Sora~\cite{sora} demonstrates the scalability and advantages of diffusion transformer architecture~\cite{dit}. Recent works such as CogVideoX~\cite{yang2024cogvideox}, Mochi1~\cite{mochi1}, and Movie Gen~\cite{polyak2024movie} have adopted the DiT architecture and achieved impressive results. In this work, we build upon the open-source model CogVideoX~\cite{yang2024cogvideox}, a DiT-based architecture, to explore attention control mechanisms for multi-prompt long video generation.
\paragraph{Long Video Generation.}
Training diffusion models on long videos often requires significant computational resources. Consequently, current video diffusion models are typically trained on videos with a limited number of frames. As a result, the quality of generated videos often degrades significantly during inference when generating longer videos. To address this problem, some works~\cite{henschel2024streamingt2v, wu2022nuwa, villegas2022phenaki, he2022latent} employ an autoregressive mechanism for long video generation. However, due to error accumulation, these methods often suffer from quality degradation after a few iterations. Alternatively, tuning-free methods~\cite{wang2023gen, qiu2023freenoise, tan2024videoinfinity, kim2024fifo, bansal2024talc} have been developed to extend off-the-shelf short-video diffusion models for generating long videos without additional training. For instance, Gen-L-Video~\cite{wang2023gen} processes long videos as short video clips with temporal overlapping during the denoising process. FreeNoise~\cite{qiu2023freenoise} explores the influence of initial noises and conducts temporal attention fusion based on the sliding window for temporal consistency. MultiDiffusion~\cite{bar2023multidiffusion} and Mimicmotion~\cite{zhang2024mimicmotion} introduces the latents blending strategy to achieve smooth transitions.
Inspired by these works, we propose a novel KV-sharing mechanism and latent blending strategy for seamless transitions between different segments without additional training.
\paragraph{Image/Video Editing with Attention Control.}
Attention control is gaining popularity due to its ability to perform zero-shot image or video editing without the need for additional data. In the realm of image editing, MasaCtrl~\cite{cao2023masactrl} enhances the existing self-attention mechanism in diffusion models by introducing mutual self-attention. This allows for querying correlated content and textures from source images, ensuring consistent and coherent edits. Prompt-to-Prompt~\cite{hertz2022prompt} utilizes cross-attention layers to control the relation between text prompts and images, which has also been adopted in many image editing works~\cite{parmar2023zero, yang2023dynamic, chen2024training}. When it comes to video editing~\cite{liao2023lovecon, liu2023videop2p, qi2023fatezero}, temporal consistency needs to be considered during attention control. Video-P2P~\cite{liu2023videop2p} extend the cross-attention control from Prompt-to-Prompt to video editing. FateZero~\cite{qi2023fatezero} fuses self-attention with a blending mask obtained by cross-attention features from the source prompt. However, all these works are designed for video-to-video translation editing with structure preservation. Differently, we aim for appearance-consistent video editing over time. Besides, none of these works explore attention control in diffusion transformers. In this paper, we are the first to analyze how the full attention in diffusion transformers could be utilized for video editing over time in multi-prompt video generation.
\section{Method}
\label{sec:method}

We tackle the challenge of zero-shot, multi-prompt longer video generation without the need for model training or optimization. This allows us to generate high-quality videos with smooth and precise inter-prompt transitions, covering various transition types~(\eg, style, camera movement, and location changes). Formally, given a pre-trained single prompt text-to-video diffusion model $\mathcal{F}$ and a sequence of $n$ prompts $\{P_1, P_2, ..., P_n\}$, the proposed $\mathrm{DiTCtrl}$ can generate a coherent longer video $\mathcal{V}_{\{1,...,n\}}$ that faithfully follows these prompts over time, which can be formulated as:
\begin{equation}
\label{eqn:mp-define}
    \mathcal{V}_{\{1,...,n\}} = \mathrm{DiTCtrl}\{\mathcal{F}(P_1),...,\mathcal{F}(P_n)\}.
\end{equation}

Below, we first give a careful analysis of MM-DiT's attention mechanisms~(Sec.~\ref{sec:dit-attn}). This analysis enables us to design a mask-guided full-attention KV-sharing mechanism for video editing over time~(Sec.~\ref{sec:kv-sharing}) in multi-prompt video generation. Finally, to ensure temporal coherence across different semantic segments, we further incorporate a latent blending strategy that enables smooth transitions in longer videos with multiple prompts (Sec.~\ref{sec:latent-blend}).

\begin{figure}[t]
    \centering
    \includegraphics[width=\columnwidth]{
     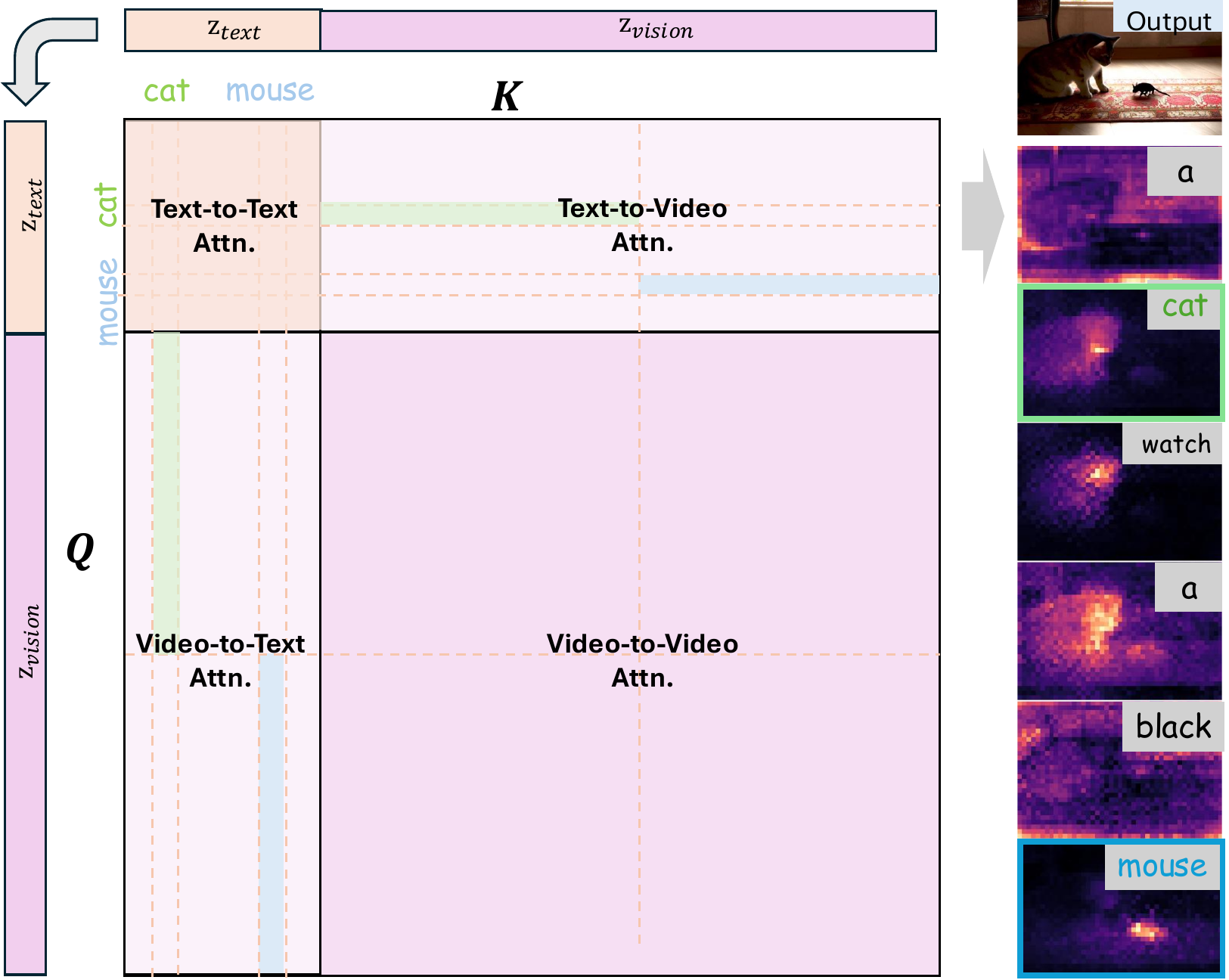}
     \vspace{-2em}
    \caption{\textbf{MM-DiT Attention Analysis}. We find the attention matrix in MM-DiT attention can be divided into four different regions. As for the prompt of \textit{`` a cat watch a black mouse''}, each text token shows a high-light response using the average of the text-to-video and video-to-text attention.
    }
    \label{fig: dit_attn_map}
\end{figure}

\subsection{MM-DiT Attention Mechanism Analysis}
\label{sec:dit-attn} 
The MM-DiT is the fundamental architecture of the current SOTA method of Text-to-image/Video models~\cite{sd3,yang2024cogvideox,mochi1,flux}, which is fundamentally distinct from prior UNet architectures since it maps text and videos into a unified sequence for attention computation. Although it has been widely utilized, the properties of its inner attention mechanism remain insufficiently explored, which restricts its applications in our multi-prompt longer video generation task. Therefore, for the first time, we conducted a comprehensive analysis of the regional attention patterns in the 3D full attention map based on the open-source video model, \ie CogVideoX~\cite{yang2024cogvideox}.

As shown in Fig.~\ref{fig: dit_attn_map}, due to the concatenation of the vision and text prompt, each attention matrix can be decomposed into four distinct regions, corresponding to different attention operations: video-to-video attention, text-to-text attention, text-to-video attention, and video-to-text attention. Below, we give the details of each region-inspired previous UNet-like structure with individual attentions~\cite{hertz2022prompt}.

\paragraph{Text-to-Video and Video-to-Text Attention.}
Previous UNet-like architectures incorporate cross-attention for video-text alignment. In MM-DiT, the text-to-video and video-to-text attention play a similar role. 
To validate its efficiency, 
we conduct a detailed analysis of the attention patterns, as illustrated in Fig.~\ref{fig: dit_attn_map}. Specifically, we compute the averaged attention values across all layers and attention heads, then extract attention values by selecting specific columns or rows corresponding to token indices in both text-to-video and video-to-text regions. These attention values are subsequently reshaped into an ${F\times H\times W}$ format, allowing us to visualize the semantic activation maps for individual frames. As demonstrated in Fig.~\ref{fig: dit_attn_map}, these visualizations show remarkable precision in token-level semantic localization, effectively capturing fine-grained relationships between textual descriptions and visual elements. This discovered capability for precise semantic control and localization provides a strong foundation for adapting established image/video editing techniques to enhance the consistency and quality of multi-prompt video generation.

\begin{figure}[t]
    \centering
    \includegraphics[width=\columnwidth]{
     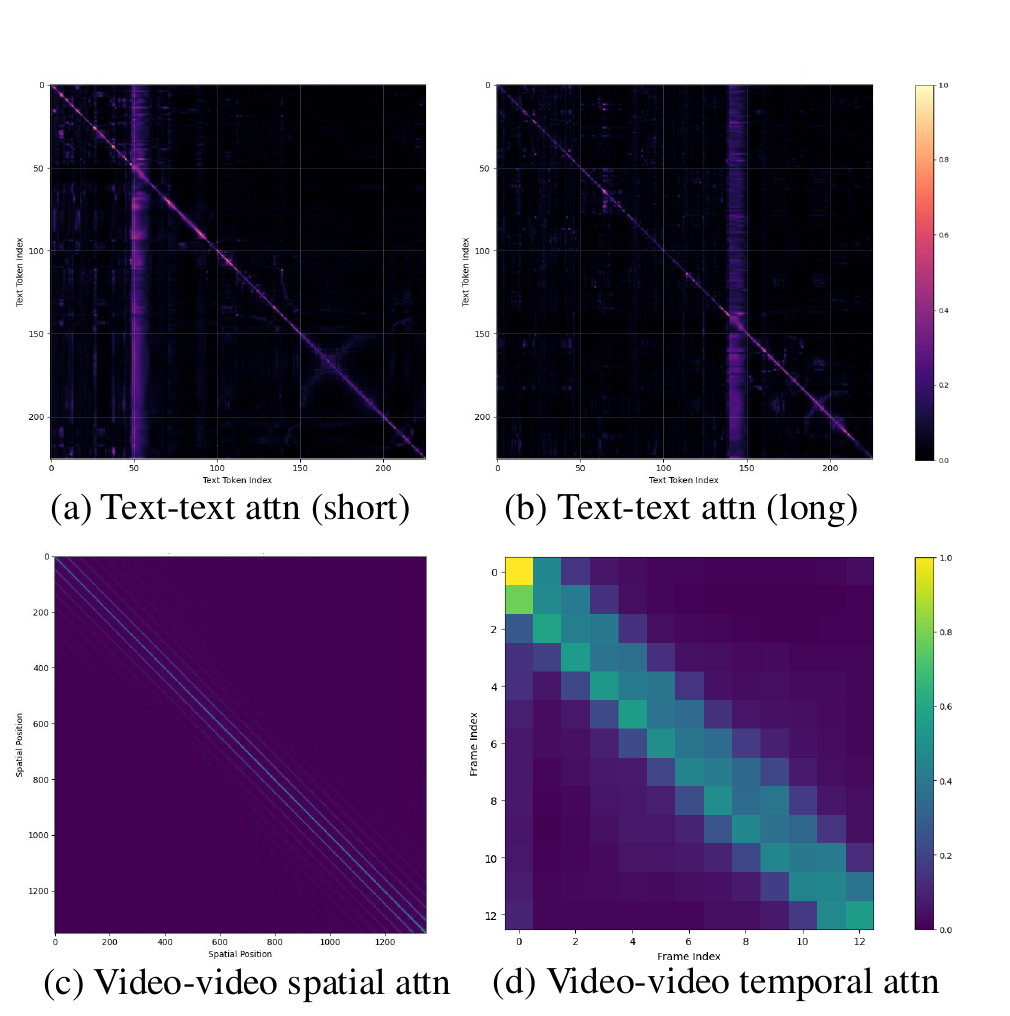}
    \caption{\textbf{MM-DiT Text-to-Text and Video-to-Video Attention Visualization}. We find that the current MM-DiT has a stronger potential to construct the individual attention in the previous UNet-like structure~\cite{chen2023videocrafter1,chen2024videocrafter2,wang2023modelscope}.
    }
    \label{fig: dit_self_attn}
\end{figure}

\textbf{Text-to-Text and Video-to-Video Attention.}
Text-to-text and video-to-video regional attention are somehow new from the respective UNet structure. As illustrated in Fig.~\ref{fig: dit_self_attn}, our analysis reveals similar patterns in both components.
In the text-to-text attention component~(Fig.~\ref{fig: dit_self_attn}(a)(b), where (a) represents the attention pattern for shorter prompts and (b) illustrates the pattern for longer prompts), we observe a prominent diagonal pattern, indicating that each text token primarily attends to its neighboring tokens. Notably, there are distinct vertical lines that shift backward as the text sequence length increases, suggesting that all tokens maintain significant attention to the special tokens at the end of the text sequence.
For the video-to-video attention component, since MMDiT flat the spatial and temporal token for 3D attention calculation, our analysis at the single-frame level reveals a distinctive diagonal pattern in spatial attention~(Fig.~\ref{fig: dit_self_attn}(c)). More significantly, when examining attention maps constructed from tokens at identical spatial positions across different frames, we also observe a pronounced diagonal pattern~(Fig.~\ref{fig: dit_self_attn}(d)). This characteristic mirrors those found in recent UNet-based video models of the spatial-attention and temporal attention, such as VideoCrafter~\cite{liu2024evalcrafter} and Lavie~\cite{wang2023lavie}, aligning with the findings reported in~\cite{lu2024freelong}. Since previous works only train the specific part of the diffusion model for more advanced control and generation, our finding provides strong evidence for these methods from MM-DiT perspectives.

Overall, the presence of these consistent diagonal patterns in the MM-DiT architecture demonstrates robust frame-to-frame correlations, which proves essential for maintaining spatial-temporal coherence and preserving motion fidelity throughout the video sequence.

\subsection{Consistent Video Generation Over Time} 
\label{sec:kv-sharing}

\begin{figure*}
    \centering
    \includegraphics[width=0.9\linewidth]{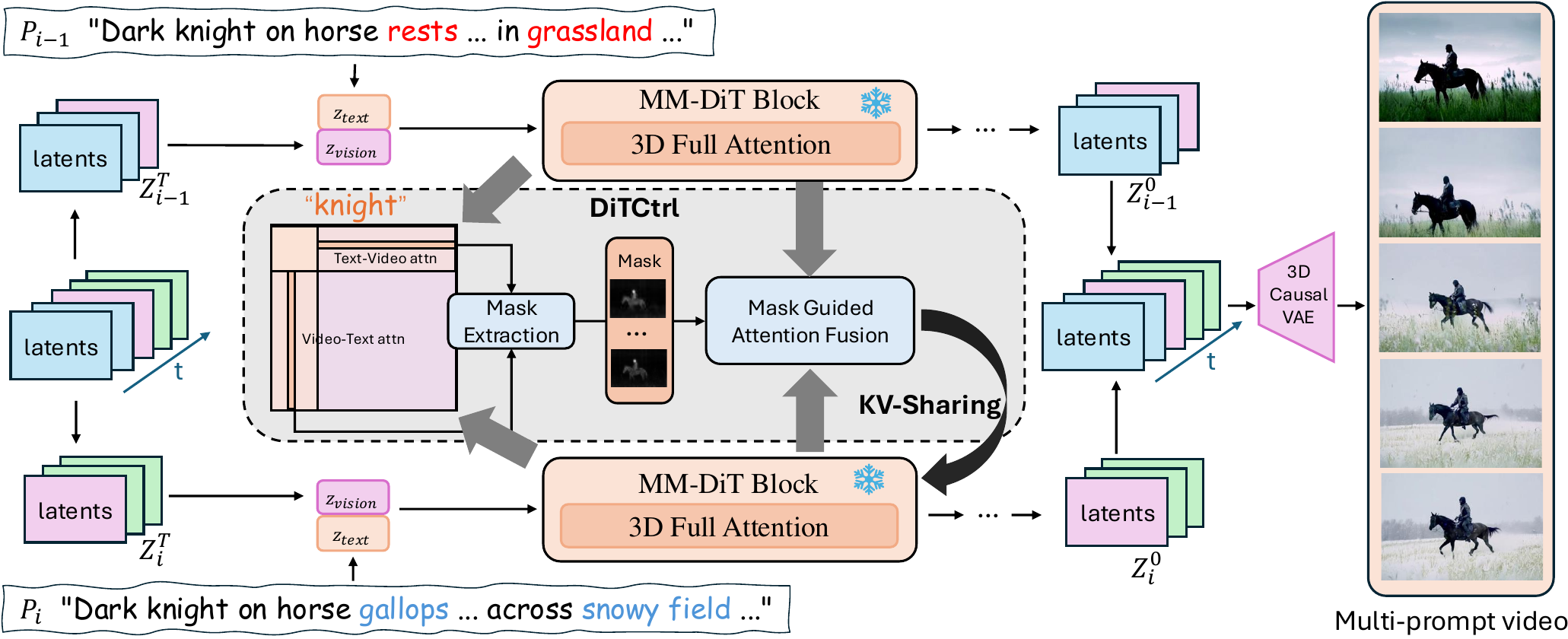}
    \caption{ \textbf{Pipeline of the proposed DiTCtrl}. Note that initial latents are assumed to be 5 frames here. The first three frames are used to generate the contents of $P_{i-1}$, and the last three frames are used to generate contents of $P_i$. The \textcolor{magenta}{pink} latent represents the overlapping frame, while the \textcolor{blue}{blue} and \textcolor{green}{green} latents are used to distinguish different prompt segments. Our method tries to synthesize content-consistent videos based on multi-prompts. The first video is synthesized with source text prompt $P_{i-1}$. During the denoising process for video synthesis, we convert the full-attention into masked-guided KV-sharing strategy to query video contents from source video $\mathcal{V}_{i-1}$, so that we can synthesize content-consistent video under the modified target prompt $P_i$. } 
    \vspace{-0.5em}
    \label{fig:main_arch}
\end{figure*}

Based on the previous analysis, 
we propose the masked-guided KV-sharing strategy for consistent video generation over time for our multi-prompt video generation task.
As shown in Fig.~\ref{fig:main_arch}, to generate the consistent video between prompt $P_{i-1}$ and prompt $P_{i}$, we utilize the intermediate attentions from the $i-1$-th and $i$-th prompt in MM-DiT to generate the attention masks of the specified foreground object (``knight'' in Fig.~\ref{fig:main_arch}). This is achieved by averaging the Text-Video/Video-Text parts of the 3D full attention across all heads and layers with the given object tokens, then thresholded to obtain binary masks $M$. Inspired by MasaCtrl~\cite{cao2023masactrl}, we leverage the masks to conduct mask-guided attention fusion based on the key and values from the prompt $P_{i - 1}$ to generate the new attention features of the prompt $P_{i}$.

We denote $M_{i-1}$ and $M_{i}$ as masks extracted for the foreground objects in videos $\mathcal{V}_{i-1}$ and $\mathcal{V}_{i}$, respectively. With these masks, we can restrict the object in $\mathcal{V}_{i}$ to query contents information only from the object region in $\mathcal{V}_{i-1}$:
\begin{align}
    \label{eq:mask_attn}
    f^l_{o} &= \text{Attention}(Q^l_{i}, K^l_{i-1}, V^l_{i-1}; M_{i-1}), \\
    f^l_{b} &= \text{Attention}(Q^l_{i}, K^l_{i-1}, V^l_{i-1}; 1 - M_{i-1}), \\
    \bar{f}^l &= f^l_{o} * M_{i} + f^l_{b} * (1 - M_{i}),
\end{align}
where $\bar{f}^l$ is the final attention output. The object regions and the background regions in the current video query the content information from corresponding restricted areas rather than all the last video features.

\if
restrict the object in $\mathcal{V}_{i}$ to query contents information only from the object region.

Then, we request the corresponding features of each model and

Inspired by~\cite{cao2023masactrl}, we propose a Full Attention KV Sharing Mechanism, which is simple but useful. The first video $\mathcal{V}_{i-1}$ is a generated one from the DiT model with text prompt $P_{i-1}$. During each denoising step $t$ of synthesizing target next video $\mathcal{V}_{i}$, we assemble the inputs of the full-attention by \textbf{1)} keeping the current Query features $Q$ unchanged, and \textbf{2)} obtaining the Key and Value features $K_{i-1}$, $V_{i-1}$ from the full-attention layer in the process of synthesizing source video clip $\mathcal{V}_{i-1}$.

However, intuitively applying such attention control across all layers and denoising steps would result in a video $\mathcal{V}_{i}$ that remains nearly identical to the previous video $\mathcal{V}_{i-1}$, failing to achieve the desired semantic transitions between prompts.

As demonstrated in Fig.~\ref{fig:intermediate_vis}, we observed phenomena in MM-DiT similar to those in UNet, where clear shapes do not form during early steps. Applying attention control too early in the process is more likely to cause semantic confusion. Additionally, the shallow layers of MM-DiT cannot effectively capture the layout and structure corresponding to the prompt, making it challenging to generate videos with the desired spatial arrangement. Therefore, we strategically apply KV-sharing to selected steps and layers, enabling both semantic transformation and enhanced temporal consistency.

\textbf{Mask-Guided KV-sharing} \label{sec:mask-guided}
We also observed the above synthesis would fail since the object and background are too similar to be confused. To tackle this problem, one feasible way is to segment the video into the foreground and the background parts and query contents only from the corresponding part. Inspired by previous work~\cite{tang2022daam,hertz2022prompt,cao2023masactrl}, the cross-attention maps correlating to the prompt tokens contain most information of the shape and structure, which is proved in the UNet-based model Stable-Diffusion. However, we found that we can utilize the regional text-video/video-text cross-attention map. As shown in~\ref{fig:main_arch},the semantic regional cross-attention maps are used to create a mask to distinguish the foreground and background in both source and target videos $\mathcal{V}_{i-1}$ and $\mathcal{V}_{i}$.

Specifically, at step $t$, we first perform a forward pass with the fixed MM-DiT backbone with prompt $P_{i-1}$ and next prompt $P_{i}$, respectively, to generate intermediate regional cross-attention maps. Then we average the cross-attention maps across all heads and layers with the same spatial resolution $H \times W$ and temporal frames $F$. The resulting cross-attention maps are denoted as ${A^c_i} \in \mathbb{R}^{F\times H\times W \times N}$, where $N$ is the number of the textual tokens. 
We then obtain the averaged cross-attention map for the token correlated to the foreground object. We denote $M_{i-1}$ and $M_{i}$ as masks extracted for the foreground objects in $\mathcal{V}_{i-1}$ and $\mathcal{V}_{i}$, respectively. With these masks, we can restrict the object in $\mathcal{V}_{i}$ to query contents information only from the object region in $\mathcal{V}_{i-1}$:

\begin{align}
    \label{eq:mask_attn}
    f^l_{o} &= \text{Attention}(Q^l_{i}, K^l_{i-1}, V^l_{i-1}; M_{i-1}), \\
    f^l_{b} &= \text{Attention}(Q^l_{i}, K^l_{i-1}, V^l_{i-1}; 1 - M_{i-1}), \\
    \bar{f}^l &= f^l_{o} * M_{i} + f^l_{b} * (1 - M_{i}),
\end{align}
where $\bar{f}^l$ is the final attention output.
The object region and the background region query the content information from corresponding restricted areas rather than all features.
\fi

\subsection{Latent Blending Strategy for Transition} \label{sec:latent-blend}

While our previous methods enable semantic consistency between adjacent video segments, achieving smooth transitions between different semantic segments still needs to be carefully designed. Thus, we propose a latent blending strategy to ensure temporal coherence across different semantic segments, inspired by recent works ~\cite{qiu2023freenoise, zhang2024mimicmotion, bar2023multidiffusion}. 

\begin{figure}[t]
    \centering
    \includegraphics[width=0.47\textwidth]{
     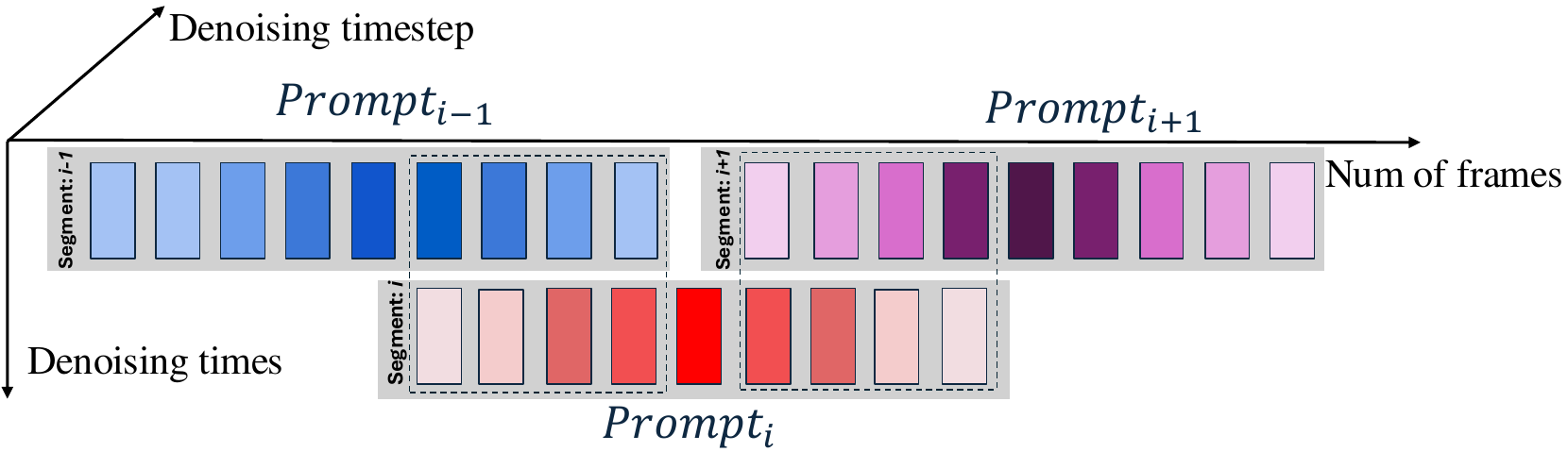}
    \caption{\textbf{Latent blending strategy} for video transition between video clips. 
    }
    \vspace{-1em}
    % \xiaodong{check all the caption}}
    \label{fig: latent_blend}
\end{figure}

As illustrated in Fig.~\ref{fig: latent_blend}, our approach introduces overlapped regions between adjacent semantic video segments. For each frame position in the overlapped region, we apply a position-dependent weight function that follows a symmetric distribution, in which frames closer to their respective segments receive higher weights while those at the boundaries receive lower weights. This weighting scheme ensures smooth transitions between different semantic contexts.

The blended latent feature $\mathbf{z}_t$ for frame $t$ is calculated as:
\begin{equation}
    \mathbf{z}_t=\frac{\sum_{i = 1}^{n}w(t_i)\cdot\mathbf{z}_{t_i}}{\sum_{i = 1}^{n}w(t_i)},
\end{equation}
\begin{equation}
    w(t_i) = \min\left(\frac{2(t_i + 0.5)}{T}, 2 - \frac{2(t_i + 0.5)}{T}\right),
\end{equation}
where $\mathbf{z}_{t_i}$ is latent feature corresponding to ${t}$-th frame in ${i}$-th latent segment, $n$ is number of overlapped segments, $T$ denotes the frames number of one latent segment and $w(t_i)$ is a position-dependent weight function.

To conclude, our approach employs the latent blending strategy and kv-sharing mechanism simultaneously during each denoising step. We process segment pairs sequentially, feeding them as one batch into the MM-DiT block for mask-guided kv-sharing (Fig.~\ref{fig:main_arch}), then blend their denoised latents progressively (Fig.~\ref{fig: latent_blend}).

\section{Experiments}
\label{sec:exp}

\begin{figure*}
    \centering
    \includegraphics[width=0.9\linewidth]{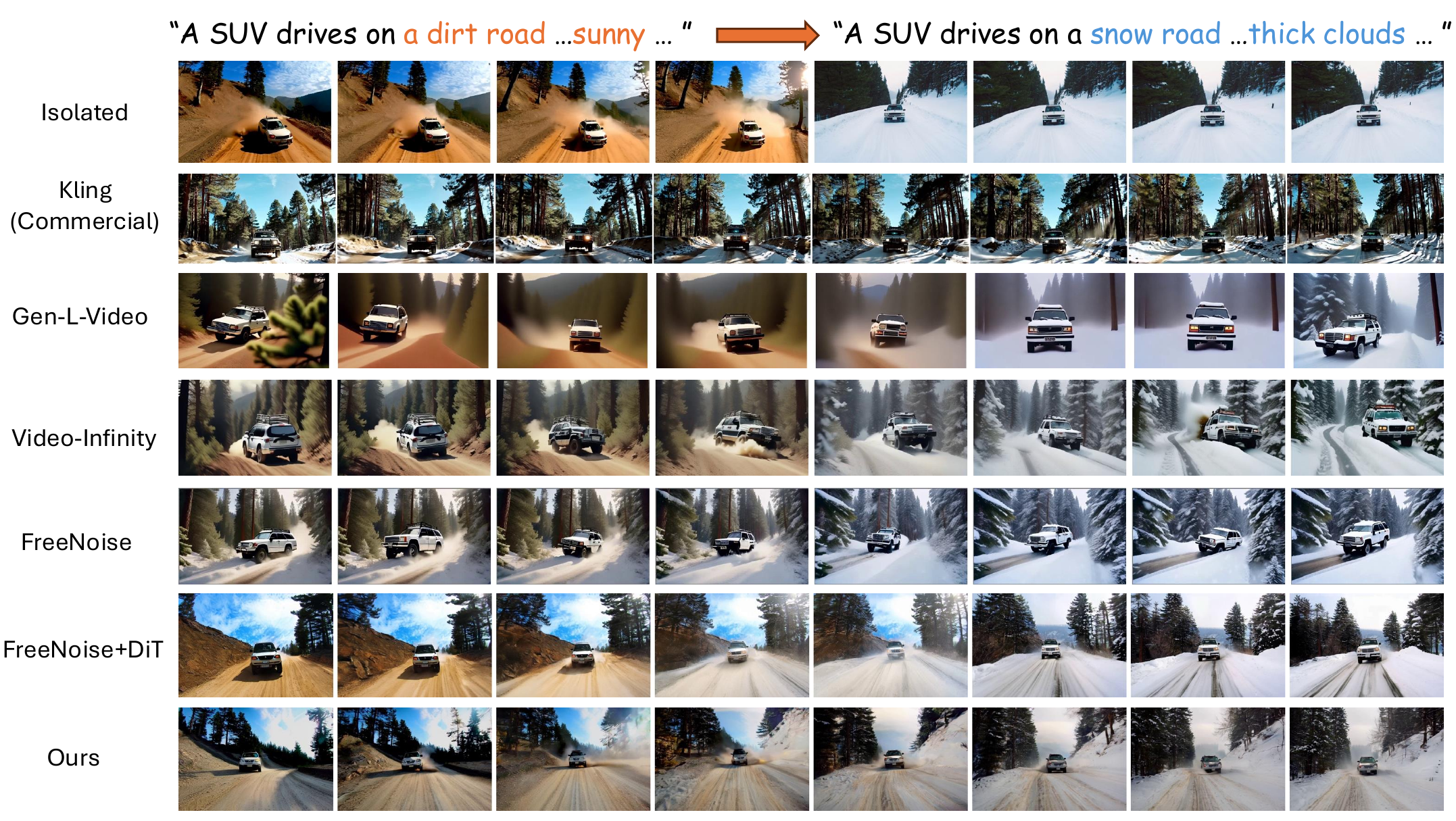}
    \vspace{-0.5em}
    \caption{Comparison of generation results across methods. Freenoise+DiT is our implementation of Freenoise on CogVideoX.} 
    \label{fig: fig_qualitative}
    \vspace{-1em}
\end{figure*}

\noindent\textbf{Setup.} 
We implement DiTCtrl based on CogVideoX-2B~\cite{yang2024cogvideox}, which is an open-source text-to-video diffusion model based on MM-DiT.
In our experimental setup, we generate multi-prompt conditioned videos, where each video clip is composed of 49 frames with a resolution of $480\times720$. The sample step is configured to 50. The kv-sharing steps are set as [2,25], and the kv-sharing layers are specified as [25,30]. For latent sampling, the number of frames is set to 13, and the overlap size is set to 6 in our experiments. All these experiments are conducted on a single NVIDIA A100 GPU. 

\noindent\textbf{Baselines.} 
We mainly compare the proposed tuning-free method to current state-of-the-art multi-prompt video generation methods~\cite{qiu2023freenoise, wang2023gen, tan2024videoinfinity} and leading commercial solutions Kling~\cite{kuaishou}. Gen-L-Video, FreeNoise, and Video-Infinity are built upon the VideoCrafter2~\cite{chen2024videocrafter2} framework. 
To ensure a fair comparison of base models, we implement FreeNoise~\cite{qiu2023freenoise} as an enhanced baseline by directly incorporating their noise rescheduling strategy into the CogVideoX framework.

\begin{figure}[t]
    \centering
    \includegraphics[width=\linewidth]{
     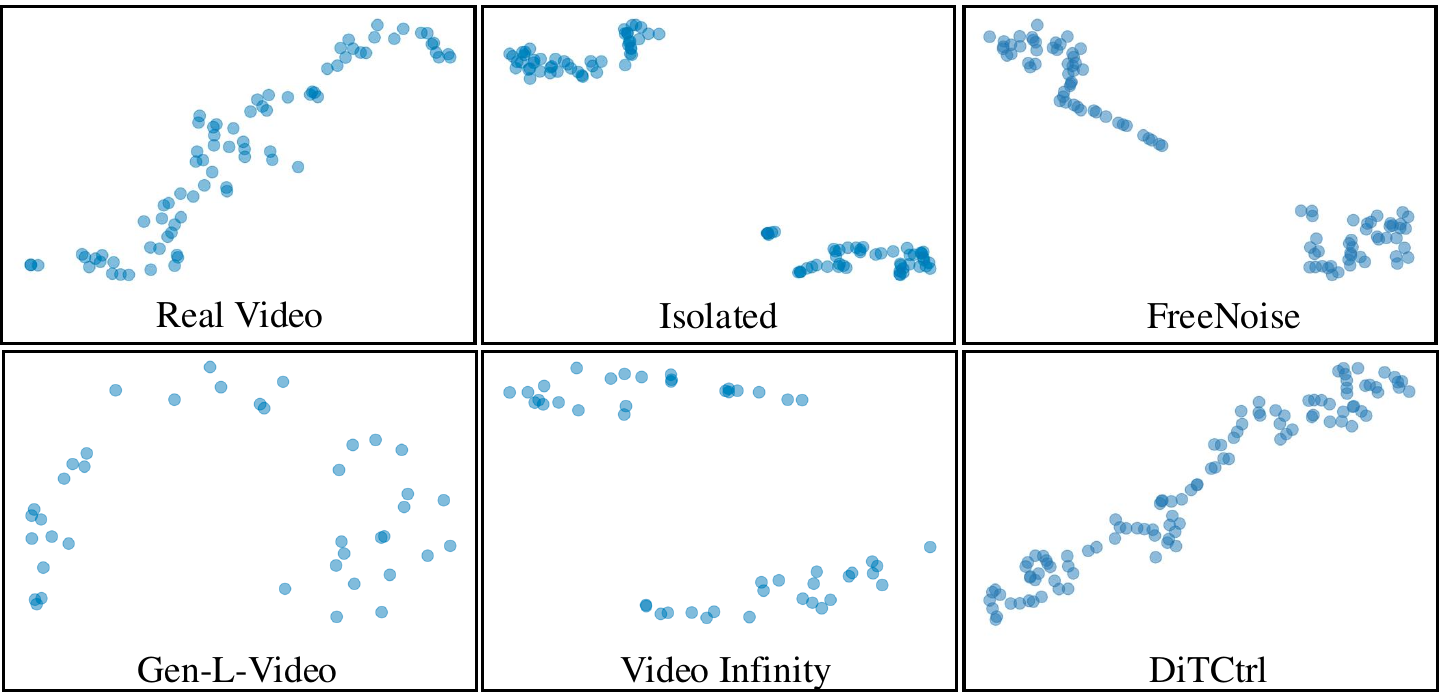}
     \vspace{-1em}
    \caption{\textbf{t-SNE visualization of CLIP embeddings}. Each point represents the CLIP embedding of a single video frame after dimensionality reduction. The visualization demonstrates that conventional multi-prompt videos form distinct clusters, while our method produces a more continuous distribution, indicating smoother semantic transitions. More details and discussions will be given in the supplementary.}
    \label{fig: tsne}
\end{figure}

\subsection{MPVBench}

MPVBench contains a diverse prompt dataset and a new metric customized for multi-prompt generation. Specifically, leveraging GPT-4, we produce $130$ long-form prompts of $10$ different transition modes. Then, for multi-prompt video generation, we observe that the distribution of the CLIP features differs between single-prompt and multi-prompt scenarios. 
As shown in Fig.~\ref{fig: tsne}, the feature points of real video (from DAVIS~\cite{PontTuset2017The2D}) follow a continuous curve, while those of two concatenated isolated videos follow two continuous curves with a breakpoint in the middle. Since the common CLIP similarity calculates the average of neighborhood similarities, the difference between real video and isolated video only occurs at the breakpoint, which becomes very small when divided by the number of frames. To address this limitation, we propose \emph{CSCV} (Clip Similarity Coefficient of Variation), a metric specifically designed to evaluate the transition smoothness of multi-prompt videos, defined as:

\begin{equation}
s_i = \mathbf{x}_i^\top \mathbf{x}_{i+1}, \quad i = 1,\ldots,n-1
\end{equation}
\begin{equation}
\text{score} = \frac{1}{1 + \lambda \cdot \frac{\sigma(s)}{\mu(s)}}\,,
\end{equation}
where $\mathbf{x}_i$ denotes frame features, $\sigma$ and $\mu$ are standard deviation and average respectively. The Coefficient of Variation $CV=\sigma(s)/\mu(s)$ describes the degree of uniformity, which can largely punish the isolated situation. The function $\frac{1}{1+\lambda (\cdot)}$ projects the score to $[0,1]$, the larger the better.
We also report the Text-Image similarity by using CLIP Similarity~\cite{hessel2021clipscore} to assess the alignment between given prompts and output video clips, and Motion smoothness from VBench~\cite{huang2024vbench} to evaluate whether the motion in the generated video is smooth, and follows the physical law of the real world.

\subsection{Qualitative Results}

The qualitative comparison with previous multi-prompt video generation methods is shown in Fig.~\ref{fig: fig_qualitative}. 
% DiTCtrl demonstrates superior performance across three critical aspects: text-to-video alignment, temporal coherence, and motion quality.
Notably, when multiple prompts are consolidated into a single-prompt describing long-term temporal changes, the generated result by Kling~\cite{kuaishou} fails to capture semantic transitions effectively, where the sun remains present while no clouds appear, which does not align with the text.
Gen-L-Video~\cite{wang2023gen} suffers from severe temporal jittering, compromising overall video quality. Video-Infinity~\cite{tan2024videoinfinity} and FreeNoise~\cite{qiu2023freenoise} both demonstrate successful scene-level semantic changes but lack physically plausible motion. For instance, in Fig.~\ref{fig: fig_qualitative}, vehicles appear to be in motion while remaining spatially fixed, which is a limitation inherent to their UNet base-model abilities. In contrast, FreeNoise+DiT leverages the DiT architecture's abilities to achieve more realistic object motion but struggles with semantic transitions, resulting in noticeable discontinuities between segments. 
Our DiTCtrl preserves the inherent capabilities of the pre-trained DiT model while addressing these limitations, enabling smooth semantic transitions and maintaining motion coherence throughout the video sequence.
For a more comprehensive evaluation, we provide additional comparisons with extensive qualitative examples in the supplementary.

\begin{table}[t]
    \setlength{\tabcolsep}{4pt} 
    \small
    \centering
    \begin{tabular}{l|ccc}
    \toprule
         \multirow{2}{*}{\textbf{Method}} & \multirow{2}{*}{\textbf{CSCV}} & \textbf{Motion} & \textbf{Text-Image} \\
         & & \textbf{smoothness} & \textbf{similarity} \\
    \hline
    Gen-L-Video                 & 67.28\% & 97.66\% & 30.60\% \\
    FreeNoise                   & 84.37\% & 97.22\% & \textbf{32.69\%} \\
    FreeNoise+DiT              & 78.74\% & 97.76\% & 30.90\% \\
    Video-Infinity             & 74.97\% & 97.31\% & 32.35\% \\
    DiTCtrl(w/o kv-sharing)    & 81.79\% & 97.35\% & 31.37\% \\
    DiTCtrl(Ours)              & \textbf{84.90\%} & \textbf{97.80\%} & 30.68\% \\
    \bottomrule
    \end{tabular}
    \vspace{-1em}
    \caption{ \textbf{Evaluation metrics.} Comparison of performance metrics for various video generation methods as benchmarked by MPVBench. Bold values represent the best performance within each group.}
    \label{tab:metrics}
\end{table}

\subsection{Quantitative Results}

% \noindent\textbf{Automatic Evaluation.}  
We conduct the automatic evaluation with our MPVBench. From Table~\ref{tab:metrics} one can see that our method achieves the highest CSCV score, demonstrating superior transition handling and overall stability in generation patterns. While FreeNoise ranks second with relatively strong stability, other methods significantly lag behind in this aspect, which is consistent with the t-SNE visualization of CLIP embedding as shown in Fig.~\ref{fig: tsne}.
In terms of motion smoothness, our approach exhibits superior performance in motion quality and consistency. Regarding Text-Image Similarity metrics, although FreeNoise and Video-Infinity achieve higher scores, this can be attributed to our method's kv-sharing mechanism, where subsequent video segments inherently learn from preceding semantic content. 

As shown in Fig.~\ref{fig: fig_qualitative}, our design choice allows the road surface to gradually transition to snowy conditions while retaining features from the previous scene. Despite potentially lower text-image alignment scores, it ensures superior semantic continuity in the sequences. In practice, this trade-off doesn't negatively impact the visual quality in multi-prompt scenarios, as demonstrated by our user study in Table~\ref{tab:user_study}.
%, with human evaluators favoring overall visual coherence and natural scene transitions.

\subsection{Human Evaluation}
% \noindent\textbf{Human Evaluation.}
We invited 28 users to evaluate five models: Gen-L-Video~\cite{wang2023gen}, Video-Infinity~\cite{tan2024videoinfinity}, FreeNoise~\cite{qiu2023freenoise}, FreeNoise+DiT and our method. We employ a Forced Ranking Scale, where items are ranked from 1 to 5, with the highest rank receiving a score of 5 and the lowest rank receiving a score of 1. Participants score each method considering overall preference, motion pattern, temporal consistency and text alignment over 16 videos generated by different scenarios. As clearly indicated in Table~\ref{tab:user_study}, generated videos from our method significantly outperform other state-of-the-art approaches in all four criteria, demonstrating superior capability in producing videos with natural semantic transitions that better align with human preferences for visual coherence and continuity.

\begin{table}[t]
    \setlength{\tabcolsep}{2pt}  
    \small
    \centering
    \begin{tabular}{l|cccc}
    \toprule
        \multirow{2}{*}{\textbf{Method}} & \textbf{Overall} & \textbf{Motion} & \textbf{Temporal} & \textbf{Text} \\
        & \textbf{preference} & \textbf{Pattern} & \textbf{Consistency} & \textbf{Alignment} \\
    \hline
    Gen-L-Video            & 1.15 & 1.14 & 1.08 & 1.25 \\
    FreeNoise              & 3.02 & 2.90 & 2.99 & 3.08 \\
    FreeNoise+DiT          & 3.81 & 3.93 & 3.75 & 3.78 \\
    Video-Infinity         & 2.90 & 2.85 & 2.91 & 2.98 \\
    DiTCtrl(Ours)          & \textbf{4.11} & \textbf{4.17} & \textbf{4.26} & \textbf{3.91} \\
    \bottomrule
    \end{tabular}
    \vspace{-1em}
    \caption{\textbf{User study}. Human evaluation of different video generation methods across multiple aspects. Scores range from 1 to 5, with higher scores indicating better performance. Bold values represent the best performance within each metric.}
    \label{tab:user_study}
\end{table}

\subsection{More Applications}
\begin{figure}[t]
    \centering
    \includegraphics[width=\columnwidth]{
     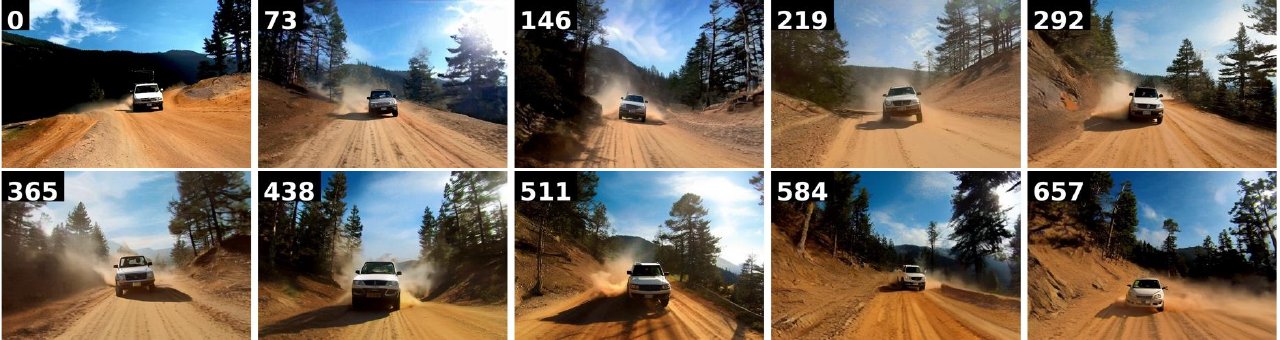}
    \vspace{-2em}
    \caption{\textbf{Single prompt longer video generation example.} 
    }
    \vspace{-1em}
    % \xiaodong{check all the caption}}
    \label{fig: single_prompt_suv}
\end{figure}

\noindent\textbf{Single-prompt Longer Video Generation.}
Our method can naturally work on single-prompt longer video generation. As illustrated in Fig.~\ref{fig: single_prompt_suv}, using the prompt “A white SUV drives on a steep dirt road”, our approach successfully generates videos that are more than 12 times longer than the original length, while maintaining consistent motion patterns and environmental coherence. 

\begin{figure}[t]
    \centering
    \includegraphics[width=\columnwidth]{
     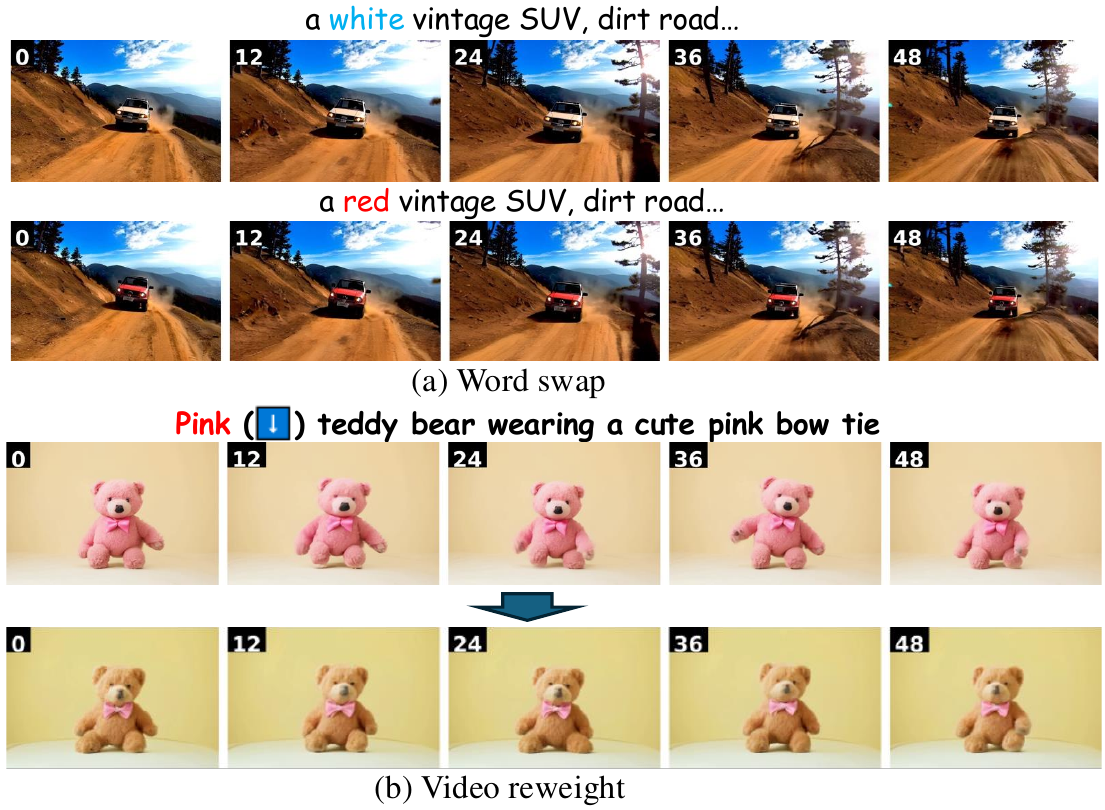}
    \vspace{-2em}
    \caption{\textbf{Video editing example.} 
    }
    \vspace{-1em}
    % \xiaodong{check all the caption}}
    \label{fig: video_edit}
\end{figure}

\noindent\textbf{Video Editing.}
We show how we use our methods to achieve video editing performance (``word swap'' and ``reweight'') in Fig.~\ref{fig: video_edit}. 
% The details are provided in Appendix~\ref{sup:application}.
\begin{itemize}
\item \textit{Word Swap}: removing our latent blending strategy of our approach DiTCtrl, we can achieve the video editing performance of Word Swap. Specifically, we just use masked-guided KV-sharing strategy to share keys and values from source prompt $P_{source}$ branch, so that we can synthesize a new video to preserve the original composition while also addressing the content of the new prompt $P_{target}$. 
\item \textit{Reweight}: Similar to prompt-to-prompt~\cite{hertz2022prompt}, through reweighting the specific columns and rows corresponding to specified token (e.g. ``pink'') in the MM-DiT's Text-Video attention and Video-Text attention, we can also achieve the video editing performance of reweight. 

\end{itemize}

\subsection{Ablation Study}

\begin{figure}[t]
    \centering
    \includegraphics[width=\columnwidth]{
     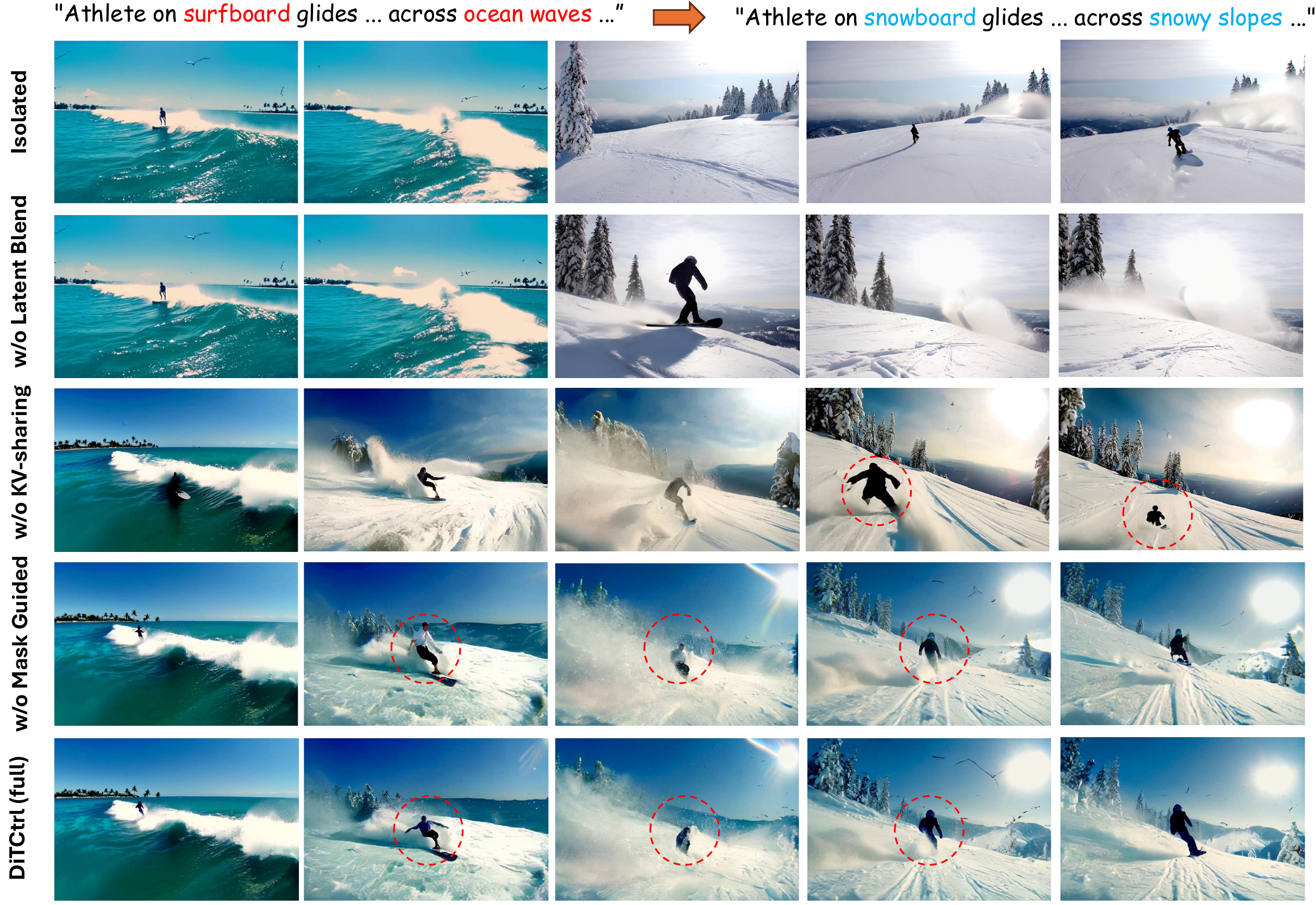}
     \vspace{-1.5em}
    \caption{\textbf{Visualization of ablation component in DiTCtrl}.}
    \vspace{-0.5em}
    \label{fig: ablation_component}
\end{figure}

We conducted ablation studies to validate the effectiveness of DiTCtrl's key components: latent blending strategy, KV-sharing mechanism, and mask-guided generation as shown in Fig.~\ref{fig: ablation_component}. 
The first row shows results that directly using text-to-video models results in abrupt scene changes and disconnected motion patterns, failing to maintain continuity in the athlete's movements from surfing to skiing. The second row demonstrates that DiTCtrl without the latent blending strategy achieves basic video editing capabilities but lacks smooth transitions between scenes. Without KV-sharing (third row), DiTCtrl exhibits unstable environmental transitions and significant motion artifacts, with inconsistent character scaling and deformed movements. Moreover, DiTCtrl without mask guidance (fourth row) improves motion coherence and transitions but struggles with object attribute confusion across different prompts and environments. On the other hand, The full DiTCtrl implementation provides the most precise control over generated content, demonstrating superior object consistency and smoother transitions between prompts while maintaining desired motion patterns. These results validate our analysis of MM-DiT's attention mechanism and its role in enabling accurate semantic control.

\section{Conclusion}
\label{sec:conclusion}

In this paper, we introduce DiTCtrl, a novel, tuning-free method for multi-prompt video generation using the MM-DiT architecture. Our pioneering analysis of MM-DiT's attention mechanism reveals similarities with the cross/self-attention blocks in UNet-like diffusion models, enabling mask-guided semantic control across prompts. With mask-guided kv-sharing mechanism and latent blending strategy, DiTCtrl ensures smooth transitions and consistent object motion between semantic segments, without extra training. We also present MPVBench, a new benchmark with diverse transition types and specialized metrics for assessing multi-prompt transitions.

\noindent\textbf{Limitation \& Future Work.} While our method demonstrates state-of-the-art performance, there remain two primary limitations. First, compared to image generation models, current open-source video generation models exhibit relatively weaker conceptual composition capabilities, occasionally resulting in attribute binding errors across different semantic segments. Second, the computational overhead of DiT-based architectures presents challenges for inference speed. These limitations suggest promising directions for future research in enhancing semantic understanding and architectural efficiency.

\noindent\textbf{Acknowledgements.} 
This work is partially supported by the National Natural Science Foundation of China (Grant No. 62306261), and The Shun Hing Institute of Advanced Engineering (SHIAE) Grant (No. 8115074).

\newpage

{\small
\bibliographystyle{ieee_fullname}
\bibliography{_main}
}

\ifarxiv \clearpage \appendix
\label{sec:appendix}

\section*{Overview}
This supplementary material presents comprehensive experimental details, qualitative analyses, and technical implementations of our work. We provide extensive evaluations across multiple aspects, including baseline comparisons, diverse application scenarios, and ablation studies. 
Note that our \href{https://onevfall.github.io/project_page/ditctrl/}{{project page}} shows many cases of our results, comparison and diverse application scenarios. The content is organized into six main sections:

\begin{itemize}
\item Section \hyperref[sup:exp]{A} details our experimental framework, including baseline implementations and model implementation details.

\item Section \hyperref[sup:eval]{B} details our evaluation, including evaluation metrics, human evaluation protocols, and TSNE visualization discussion.

\item Section \hyperref[sup:exp_result]{C} showcases comprehensive qualitative results across diverse domains, featuring detailed comparisons with state-of-the-art models and demonstrating the versatility of our approach.

\item Section \hyperref[sup:application]{D} explores various applications, including single-prompt video generation and advanced editing capabilities such as attention reweighting and word swap techniques.

\item Section \hyperref[sup:generator]{E} presents the usage of \textit{prompt generator}, including full descriptions used to generate individual prompts.

\item Section \hyperref[sup:ablation]{F} presents comprehensive ablation studies, including both quantitative evaluations and qualitative analyses of the masking mechanism.
\item Section \hyperref[sup:infer_time]{G} discuss the inference time of alternative methods.
\end{itemize}

\section{Implementation Details}
\label{sup:exp}
\noindent\textbf{Details.} 
We implement DiTCtrl based on CogVideoX-2B~\cite{yang2024cogvideox}, which is a state-of-the-art open-source text-to-video diffusion model based on MM-DiT. The hyperparameters and implementation details are shown in \cref{tab:hyperparameters}. 
\begin{table}[h!]
    \centering
    \caption{
        Hyperparameters of DiTCtrl. 
    }
    
    \begin{tabular}{l|c}
        \toprule
        \textbf{Hyperparameters} & \\ 
        \midrule
        base model & CogVideoX-2B \\
        sampler & VPSDEDPMPP2MSampler \\
        sample step & 50 \\
        guidance scale & 6 \\
        resolution & $480 \times 720$ \\
        video frames & 49 \\
        latent num frames & 13 \\
        overlap size & 6 \\
        kv-sharing steps & [2,25] \\
        kv-sharing layers & [25,30] \\
        threshold & 0.3 \\
        $\lambda$ of CSCV & 10\\
        \bottomrule
    \end{tabular}
    
    \label{tab:hyperparameters}
\end{table}

\noindent\textbf{Baselines.} 
In experiments of our main paper, we comprehensively compare our method with previous state-of-the-art methods, including commercial and open-source techniques. We offer more details of the baselines that we use here:
\begin{itemize}
    \item \textbf{Kling~\cite{kuaishou}}: Kling is leading closed-source commercial solutions developed by Kuaishou Technology. It can generate videos of 6s lengths, but it can only input single-prompt, so we input a single prompt describing long-term temporal changes. We use the Kling1.5 model for our visualization comparison.
    \item \textbf{Gen-L-Video~\cite{wang2023gen}}: Gen-L-Video processes long videos as short video clips with temporal overlapping during the denoising process. We use the VideoCrafter2~\cite{chen2024videocrafter2} as the base model.
    \item \textbf{FreeNoise~\cite{qiu2023freenoise}}: FreeNoise reschedules the initial noise sequence and conducts temporal attention fusion based on the sliding window for temporal consistency. We use the VideoCrafter2~\cite{chen2024videocrafter2} as the base model.

    \item \textbf{Video-Infinity~\cite{tan2024videoinfinity}}: Video-Infinity scales up long video generation via distributed inference. We use the VideoCrafter2~\cite{chen2024videocrafter2} as the base model.
    \item \textbf{FreeNoise+DiT}: This is an enhanced baseline by directly incorporating FreeNoise's noise rescheduling strategy into the CogVideoX~\cite{yang2024cogvideox} framework.
\end{itemize}
For a fair comparison, all baseline methods should be aligned to use the same ratio stride.
Since CogVideoX-2B has 13 latent frames, we used overlap frame 6 in our paper which is approximately 1/2 stride of the total frames ($6/13 \approx 1/2$). Other baseline methods also use this setting of same stride ratio.

\noindent\textbf{Mask-guided Implementation Details.} 
We show how mask extracted from MM-DiT attention map is utilized for mask-guided KV-sharing strategy in Fig.~\ref{fig:mask_guided_strategy}, to generate consistent video over time for multi-prompt video generation task. 

Specifically, Fig.~\ref{fig:mask_guided_strategy} illustrates our approach to generating temporally consistent videos in multi-prompt video generation tasks. When computing attention for the $P_{i}$ branch latent, we utilize attention maps from both $P_{i-1}$ and $P_{i}$ branches. Specifically, we extract content from the Text-video and Video-text attention regions of their attention maps. By focusing on specified tokens (e.g., ``a running horse"), we obtain and average the corresponding regional values to generate semantic mask maps. These maps are then binarized through thresholding to create foreground-background segmentation masks $M_{i-1}$ and $M_i$.

Then, we leverage $M_{i-1}$ to guide the computation of KV-sharing attention maps (calculating attention between $Q_i$ and $K_{i-1}$, $V_{i-1}$), resulting in foreground-focused attention outputs $F_{fore}$ and $F_{back}$. The final fusion is achieved through $M_i$ as follows:

\begin{equation}
F_{fusion} = F_{fore} * M_i + F_{back} * (1-M_i)
\end{equation}

This mask-guided approach ensures semantic consistency while maintaining smooth transitions between different prompts.

\begin{figure*}
    \centering
    \includegraphics[width=0.95\linewidth]{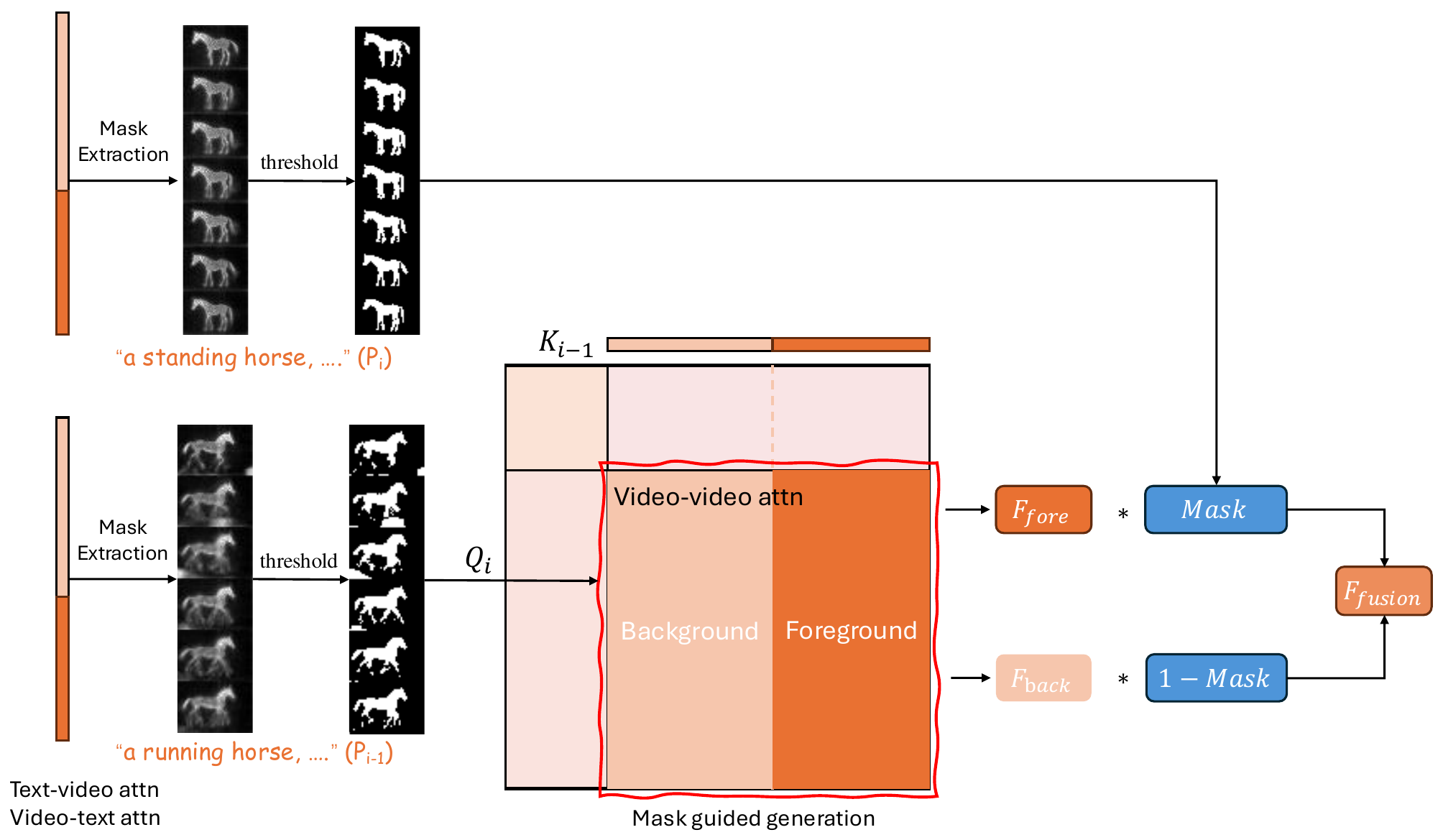}
    \vspace{-0.5em}
    \caption{Mask-guided KV-sharing details.} 
    \label{fig:mask_guided_strategy}
    \vspace{-1em}
\end{figure*}

\section{Evaluation details}
\label{sup:eval}

\noindent\textbf{MPVBench.} 
We introduces a new benchmark MPVBench, which is specified designed for multi-prompt video generation task. MPVBench contains a diverse prompt dataset and a new metric customized for multi-prompt generation. Specifically, leveraging GPT-4, we produce $130$ long-form prompts of $10$ different transition modes (background transition, subject transition, camera transition, style transition, lighting transition, location transition, speed transition, emotion transition, clothing transition, action transition). The instruction of prompt generator is provided in Fig.~\ref{fig:prompt_gen_instruction}.

\noindent\textbf{Automatic evaluation.} 
For automatic evaluation, we generate videos using 130 prompts from our MPVBench, with three random seeds set. Then, we evaluate the generated video by three metrics: CSCV (Clip Similarity Coefficient of Variation), Motion Smoothness, Text-Image Similarity.

\noindent\textbf{Human evaluation.} 
In our user study, we combined our generated videos with those produced by four other baseline methods. We asked a total of 28 participants to evaluate the videos across four dimensions: overall preference, motion pattern, temporal consistency, and text alignment. Specifically, we asked all participants to rank the results of these methods for each of the following questions, and assigned a score from 1 (lowest quality) to 5 (highest quality) for these five methods:
\begin{itemize}
    \item \textbf{Overall Preference}: ``\textit{Please rank the overall video preference.}" This metric evaluates participants' comprehensive assessment of the generated videos.
    
    \item \textbf{Motion Pattern}: ``\textit{How natural and realistic are the motion in the video?}" This evaluates whether the motion of objects in the generated video appears physically plausible and natural, such as whether vehicles drive realistically, animals move naturally, or human actions appear authentic.
    \item \textbf{Temporal Consistency}: ``\textit{How smoothly does the video content transition across different frames?}" This metric evaluates the temporal coherence of the generated video, focusing on whether the transitions between consecutive frames are natural and continuous, without abrupt changes or visual artifacts. It measures the video's ability to maintain visual continuity throughout its duration.
    \item \textbf{Text Alignment}: ``\textit{To what extent does the video content match the given text descriptions?}" This assesses the semantic fidelity between the generated visual content and the input text prompts, examining whether the video accurately captures and visualizes the key elements and actions described in the prompts. It measures how well the visual narrative aligns with the intended textual description.
\end{itemize}

\noindent\textbf{t-SNE Visualization discussion.} 
In the justification for the proposed CSCV metric, which evaluates the transition smoothness, We found that t-SNE visualizations of real videos from existing datasets have similar continuous trajectories due to semantic continuity. Therefore, we just present one representative case, the t-SNE of video embeddings for real videos. The selected real video in Fig.7 of main paper is the classic car video from DAVIS~\cite{PontTuset2017The2D}. The car video frames are shown in Fig.\ref{fig: real_video}.

We also show more t-SNE visualization of our comparison cases in Fig.~\ref{fig:fig_qualitative2} and Fig.~\ref{fig:fig_qualitative3}. 
Even when processing multi-prompt videos, our method generates continuous trajectories that are comparable to those in real videos. This showcases the exceptional transition handling capabilities and overall stability of the videos produced by DiTCtrl.

\begin{figure}[t]
    \centering
    \includegraphics[width=\columnwidth]{
     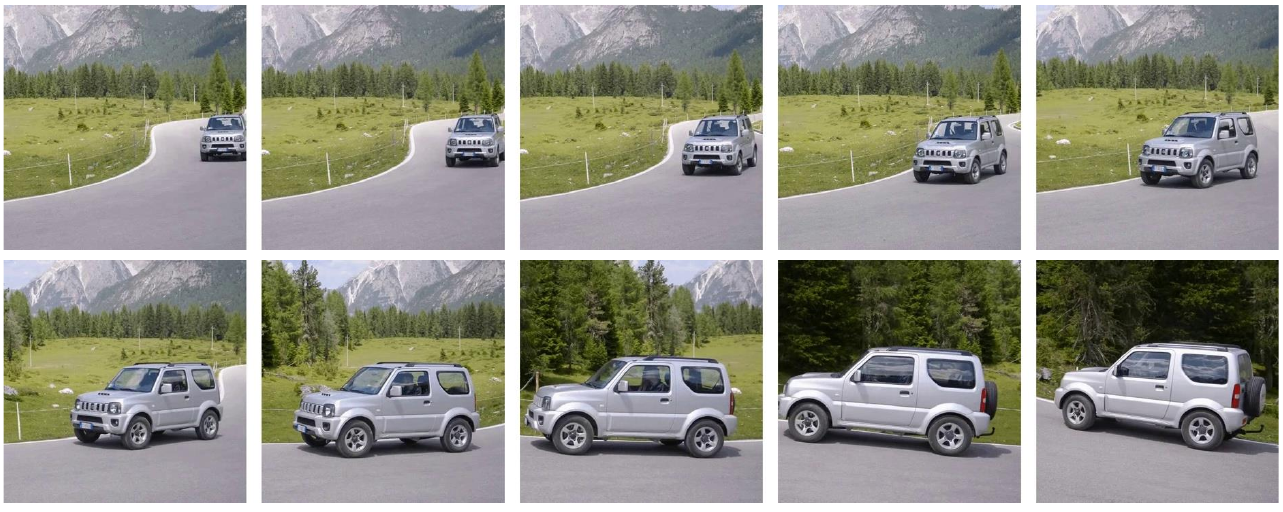}
    \vspace{-2em}
    \caption{\textbf{real video example from DAVIS~\cite{PontTuset2017The2D}.} 
    }
    \vspace{-1em}
    % \xiaodong{check all the caption}}
    \label{fig: real_video}
\end{figure}

\begin{figure}[t]
    \centering
    \includegraphics[width=\columnwidth]{
     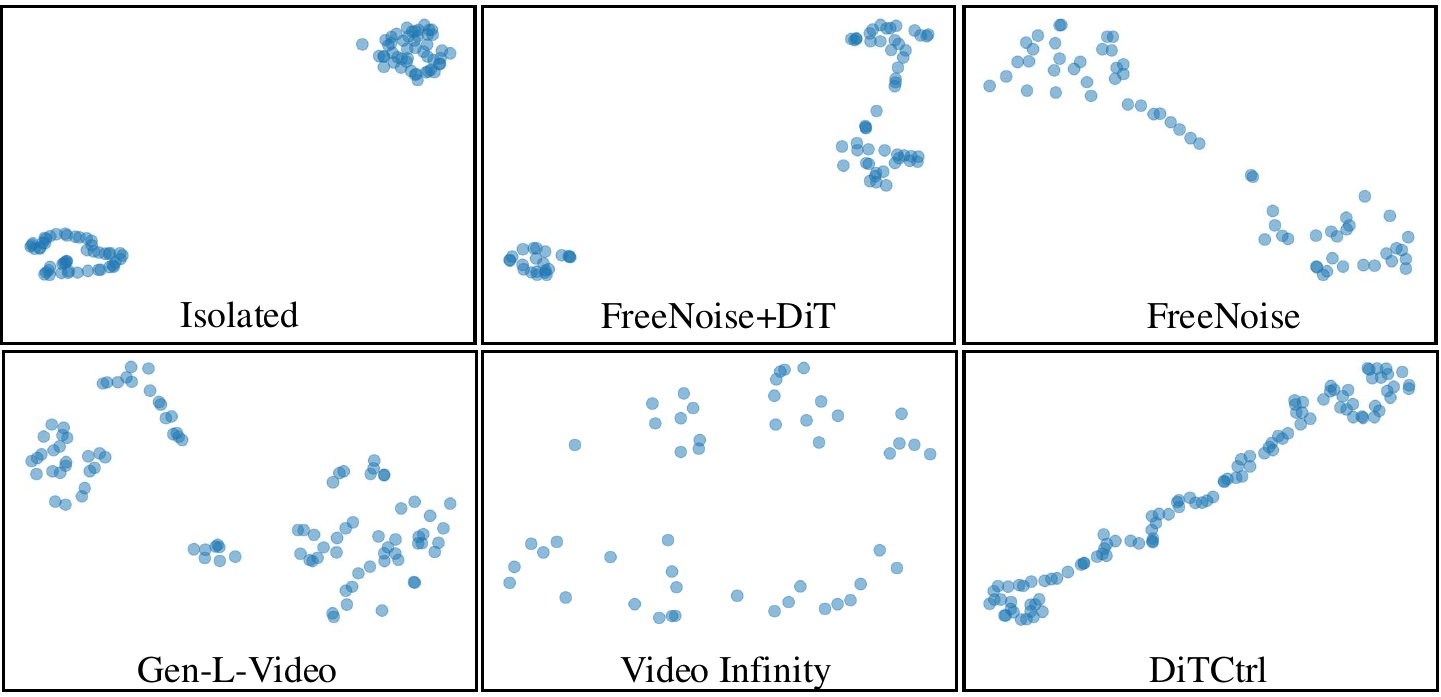}
    \vspace{-2em}
    \caption{\textbf{t-SNE Visualization} of Fig.~\ref{fig:fig_qualitative2}
    }
    \vspace{-1em}
    % \xiaodong{check all the caption}}
    \label{fig: tsne_dark_knight}
\end{figure}

\begin{figure}[t]
    \centering
    \includegraphics[width=\columnwidth]{
     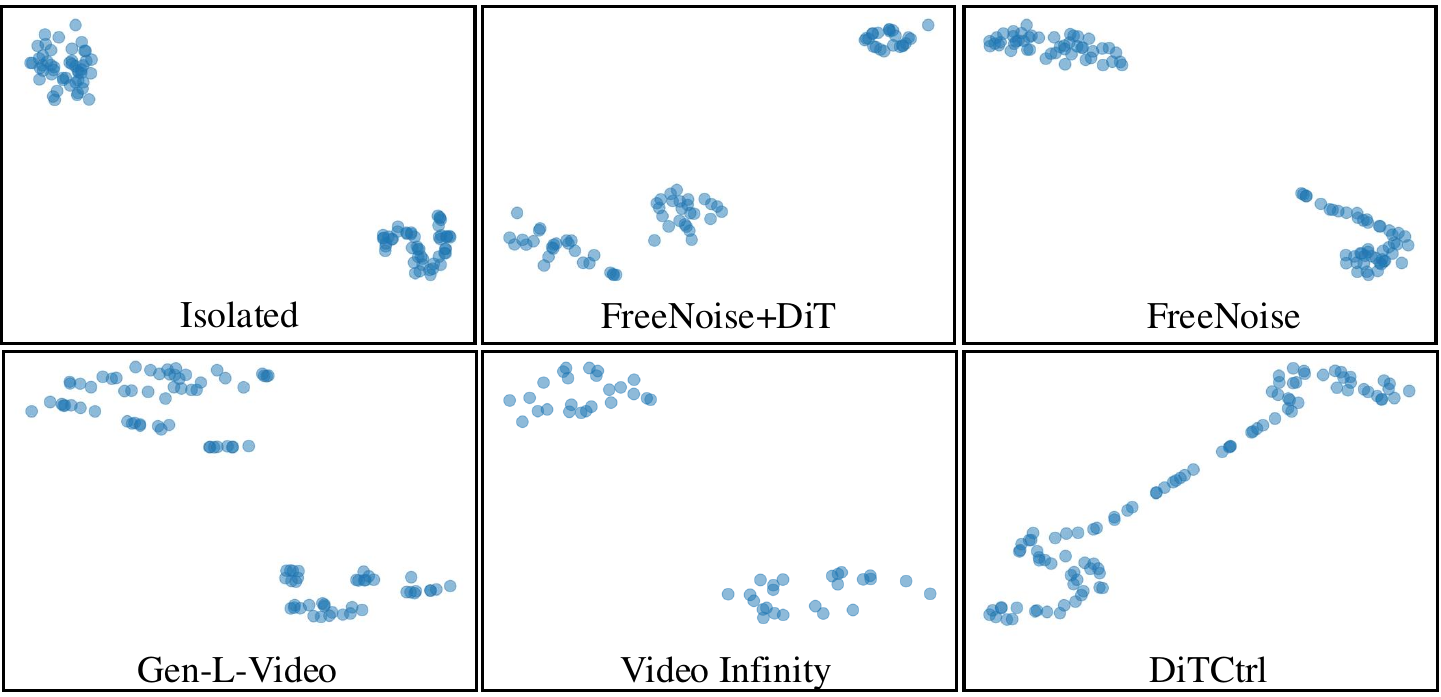}
    \vspace{-2em}
    \caption{\textbf{t-SNE Visualization} of Fig.~\ref{fig:fig_qualitative3}
    }
    \vspace{-1em}
    % \xiaodong{check all the caption}}
    \label{fig: tsne_street}
\end{figure}

\section{More Qualitative Results}
\label{sup:exp_result}
More results are provided in Fig.~\ref{fig:more_result_1} and Fig.~\ref{fig:more_result_2}. Our method DiTCtrl can generate multi-prompt videos with good temporal consistency and strong prompt-following capabilities, demonstrating cinematographic-style transitions in depicting the boy's riding sequence. 
We also give more qualitative comparisons with state-of-the-art multi-prompt video generation methods~\cite{qiu2023freenoise, wang2023gen, tan2024videoinfinity}, our reproduced FreeNoise+DiT, and leading commercial solutions Kling~\cite{kuaishou}. We show the \textit{motion transition} case, and \textit{background transition} case in Fig.~\ref{fig:fig_qualitative2} and Fig.~\ref{fig:fig_qualitative3}. 
Our comparative analysis reveals distinct characteristics and limitations of existing approaches. Gen-L-Video~\cite{wang2023gen} suffers from severe temporal jittering, compromising overall video quality. Video-Infinity~\cite{tan2024videoinfinity} and FreeNoise~\cite{qiu2023freenoise} both demonstrate successful scene-level semantic changes but lack physically plausible motion. For instance, in Fig.~\ref{fig:fig_qualitative2}, dark knight appear to be in motion while remaining spatially fixed, which is a limitation inherent to their UNet-based abilities. In contrast, FreeNoise+DiT leverages the DiT architecture's abilities to achieve more realistic object motion but struggles with semantic transitions, resulting in noticeable discontinuities between segments. 
Our proposed DiTCtrl method preserves the inherent capabilities of the pre-trained DiT model while addressing these limitations, enabling smooth semantic transitions and maintaining motion coherence throughout the video sequence.
More comparison of visualization case and our results are shown in our \href{https://onevfall.github.io/project_page/ditctrl/}{{project page}}.

\section{Applications}
\label{sup:application}

Based on our exhaustive analysis and exploration of attention control in MM-DiT architecture, our method could be applied to other tasks like single prompt longer video generation and video editing and achieves promising results.

\subsection{Single-prompt Longer Video Generation}
Although our primary objective is to address multi-prompt video generation, we discover that our method demonstrates remarkable effectiveness in single-prompt longer video generation as well.
Our method can naturally work on single-prompt longer video generation. As illustrated in Fig.~\ref{fig:single_prompt_sup}, our approach successfully generates longer videos, while maintaining consistent motion patterns and environmental coherence.

\subsection{Video Editing}
In this work, we conduct an in-depth analysis of MM-DiT's attention maps, which can be categorized into four components: Text-to-Video and Video-to-Text Attention, Text-to-Text and Video-to-Video Attention. Through our analysis of Text-to-Video and Video-to-Text Attention, we observe that semantic maps can be obtained by specifying token indices, suggesting potential for semantic control. We have emphasized the use of extracted foreground-background segmentation semantic maps to guide video generation, effectively preventing semantic confusion between foreground and background elements. In this section, we demonstrate video editing capabilities through two approaches: \textit{Reweight} and \textit{Word Swap}.

\noindent\textbf{Attention Re-weighting.} As illustrated in Fig.~\ref{fig:reweight}, we can achieve semantic enhancement or attenuation by increasing or decreasing the values in rows or columns corresponding to token $j$ in the Text-to-Video and Video-to-Text Attention maps. In Fig.~\ref{fig:reweight}~(a), we demonstrate semantic attenuation by reducing Text-Video Attention values in the row and Video-Text Attention values in the column corresponding to ``pink". In Fig.~\ref{fig:reweight}~(b), we achieve semantic enhancement by increasing Text-Video Attention values in the row and Video-Text Attention values in the column corresponding to ``snowy". These results validate the semantic control capabilities of Text-Video and Video-Text Attention in MM-DiT.

\noindent\textbf{Word Swap.} Building upon the concept introduced in Prompt-to-prompt~\cite{hertz2022prompt}, this approach allows users to swap tokens in the original prompt with alternatives (e.g., changing P =``a large bear" to ``a large lion"). The primary challenge lies in maintaining the original composition while accurately reflecting the content of the modified prompt. Our DiTCtrl method incorporates KV-sharing, similar to the word swap mechanism in~\cite{hertz2022prompt}, where we share key-value pairs from the previous prompt to compute the corresponding video for the subsequent prompt across selected layers and steps. Specifically, DiTCtrl (without latent-blending strategy) enables token-replacement video editing while ensuring consistency in other content elements, as demonstrated in Fig.~\ref{fig: word_swap}. This implementation validates the feasibility of prompt-to-prompt-style video editing within the MM-DiT architecture.

\section{Prompt Generator}
\label{sup:generator}
In this section, we provide additional information of the prompt generator that is described in our main paper. We use GPT4 for longer multi-prompt generation, our prompts are shown in Fig.~\ref{fig:prompt_gen_instruction}. This figure shows the generation process of "background transition", and we generate 10 different transition modes (background transition, subject transition, camera transition, style transition, lighting transition, location transition, speed transition, emotion transition, clothing transition, action transition). 
% The detailed prompts will be provided in the the \textit{prompts.txt} in the attached supplementary.

\section{Ablation Study}
\label{sup:ablation}

\subsection{Quantitative Results of Components}
As shown in Tab.~\ref{tab:ablation_metric}, our latent blending strategy (second row) demonstrates superior video consistency compared to isolated clips (first row), as evidenced by higher CSCV scores - our proposed metric for evaluating multi-prompt transition smoothness. Furthermore, our KV-Sharing mechanism further improves the CSCV value, achieving enhanced stability. The mask-guided approach(fourth row) and its unmasked counterpart(third row) report comparable scores, suggesting that the contribution of masking foreground object to overall frame transition smoothness is modest. However, our qualitative analysis in Section~\ref{sup:mask_guided_case} reveals that the mask-guided method yields superior visual results.

Additionally, in our evaluation of motion smoothness, our full method (DiTCtrl) achieves optimal performance. Regarding the Text-Image similarity metric, we observe a slight expected decrease with our approach. This is attributable to our methodology where the latent representation of the latter video segments incorporates keys and values from preceding segments to maintain consistency. This inherently introduces semantic information from previous segments, marginally reducing the current segment's alignment with its corresponding text prompt. However, this trade-off is justified as our method achieves stable transitions and effectively conveys both semantic elements, resulting in higher user study scores as shown in Tab.~\ref{tab:user_study}.

\begin{table}[t]
    \setlength{\tabcolsep}{4pt}
    \small
    \centering
    \begin{tabular}{l|cccc}
    \toprule
         \multirow{2}{*}{\textbf{Method}} & \multirow{2}{*}{\textbf{CSCV}} & \textbf{Motion} & \textbf{Text-Image} \\
         & & \textbf{smoothness} & \textbf{similarity} \\
    \hline
    Isolated                   & 72.37\% & 97.78\% & \textbf{32.05\%} \\
    DiTCtrl(w/o kv-sharing)    & 81.79\% & 97.35\% & 31.37\% \\
    DiTCtrl(w/o mask-guided)   & \textbf{84.92\%} & 97.76\% & 30.66\% \\
    DiTCtrl(full)              & 84.90\% & \textbf{97.80\%} & 30.68\% \\
    \bottomrule
    \end{tabular}
    \vspace{-1em}
    \caption{Comparison of metrics for ablation.}
    \label{tab:ablation_metric}
\end{table}

\begin{figure*}
    \centering
    \includegraphics[width=0.95\linewidth]{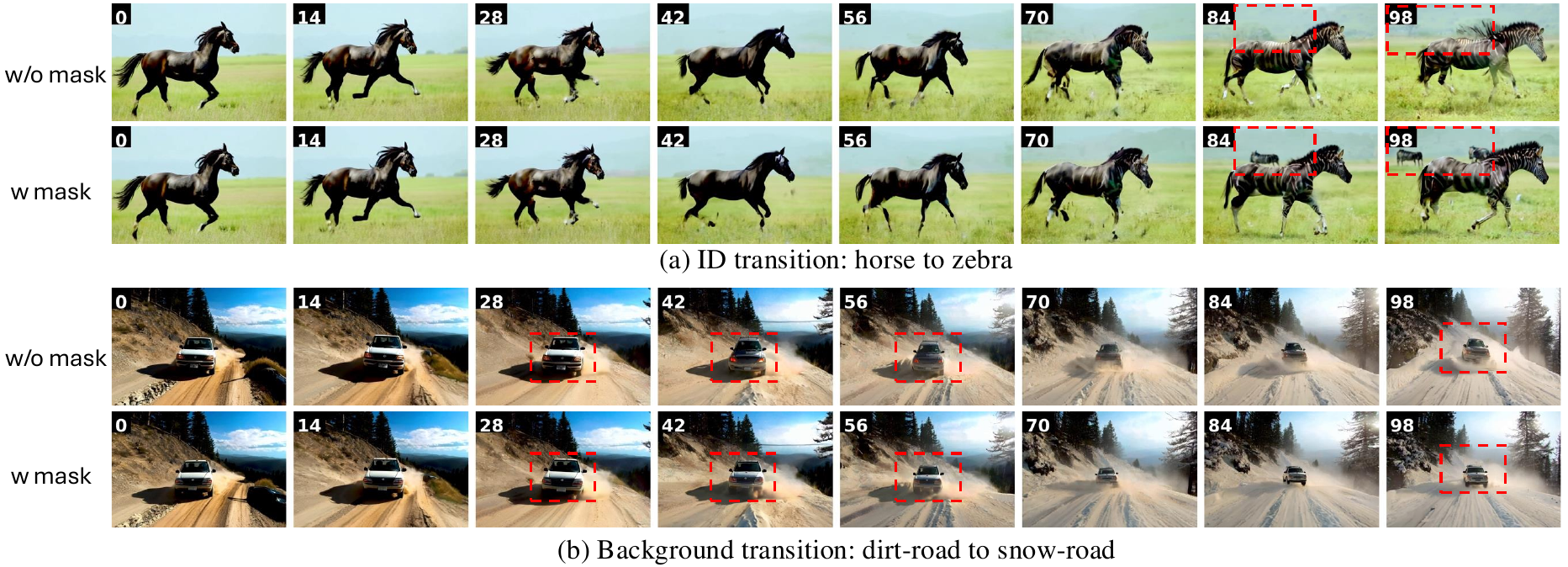}
    \vspace{-0.5em}
    \caption{Ablation study of mask-guided KV-sharing results. First row shows our model without mask-guided KV-sharing, while the second row demonstrates our full model with mask-guided KV-sharing. The prompt for (a) transitions from ``A powerful horse gallops across a field...'' to ``A striking zebra leads its herd across the field...''. The prompt for (b) evolves from ``A white SUV drives a dirt road...'' to ``A white SUV powers through snow...''} 
    \label{fig:ablation_mask_auto}
    \vspace{-1em}
\end{figure*}

\subsection{Mask-guided Generation Analysis}
\label{sup:mask_guided_case}

We present comparative results in Fig.~\ref{fig:ablation_mask_auto} to demonstrate the effectiveness of our mask-guided KV-sharing strategy. In Fig.~\ref{fig:ablation_mask_auto}~(a), while the first prompt describes a single horse, the second prompt emphasizes a zebra leading its herd. Without mask-guided KV-sharing (first row), we observe that the model fails to properly generate the zebra herd and exhibits background inconsistencies. In contrast, our full model with mask-guided KV-sharing (second row) successfully maintains scene coherence while incorporating the herd elements.

Similarly, in Fig.~\ref{fig:ablation_mask_auto}~(b), the transition sequence in the first row (without mask-guided KV-sharing) shows notable deformations in the vehicle's appearance, including undesired color variations. The second row, implementing our mask-guided approach, better preserves the vehicle's original appearance, color, and shape throughout the transition. These results validate both the effectiveness of our mask-guided approach and the feasibility of leveraging semantic maps extracted from MM-DiT's Text-Video and Video-Text Attention for application in Video-Video Attention.

\section{Inference Time}
\label{sup:infer_time}

We present a comparison of the inference times on a single A100 GPU, with the variation based on the number of prompts (N). For a fair assessment, when 2 prompts are input, each method is tasked to generate approximately 100 frames. When the number of prompts increases to 3, the generation target is set at approximately 150 frames. 
As depicted in Table~\ref{tab:time}, our method (without mask) demonstrates competitive efficiency in terms of elapsed time, and also achieves satisfactory video transition effects. When the mask-guided approach is further employed, it yields even more superior visual outcomes. Despite the sixfold increase in runtime,  the method remains Pareto optimal.

\begin{table*}[h!]
    \setlength{\tabcolsep}{4pt}  
    \setlength{\textfloatsep}{0pt}  
    \setlength{\abovecaptionskip}{0pt}  
    \setlength{\belowcaptionskip}{0pt}  
    \centering
    \begin{tabular}{l|cccccc}
    \toprule
         & Gen-L-Video & FreeNoise & FreeNoise+DiT & Video-Infinity & Ours(w/o mask) & Ours(w/ mask) \\
    \midrule
    N=2 & 9.1min & 6.1min & 5.3min & 1.2min (2 gpu) & 5.3min & $\sim$39min \\
    N=3 & 13.6min & 9.2min & 10.6min & 1.2min (3 gpu) & 10.6min & $\sim$78min \\
    \bottomrule
    \end{tabular}
    \caption{Inference time comparison with the number of prompts N}
    \label{tab:time}
\end{table*}

\begin{figure*}
    \centering
    \includegraphics[width=0.95\linewidth]{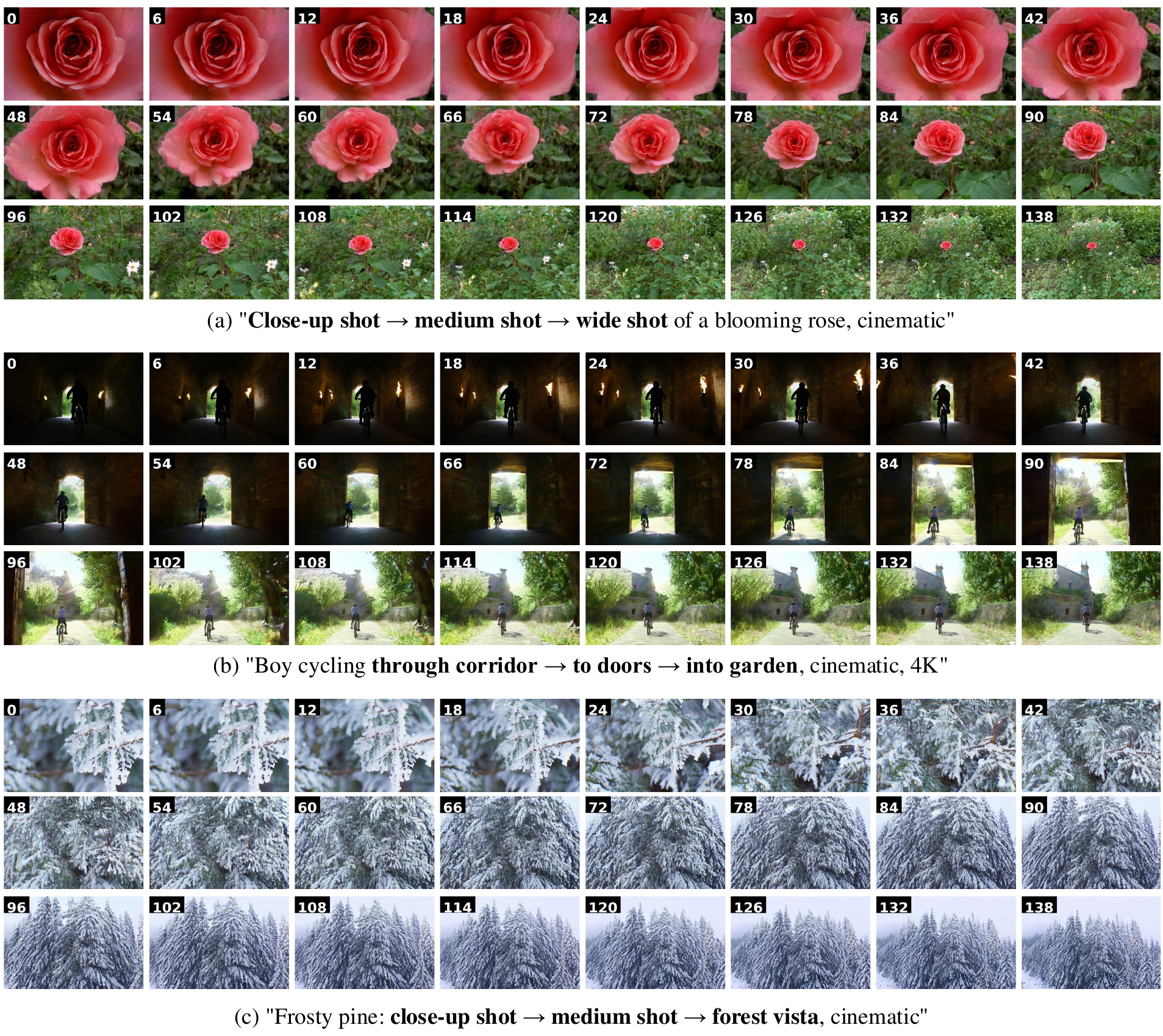}
    \vspace{-0.5em}
    \caption{More multi-prompt results} 
    \label{fig:more_result_1}
    \vspace{-1em}
\end{figure*}

\begin{figure*}
    \centering
    \includegraphics[width=0.95\linewidth]{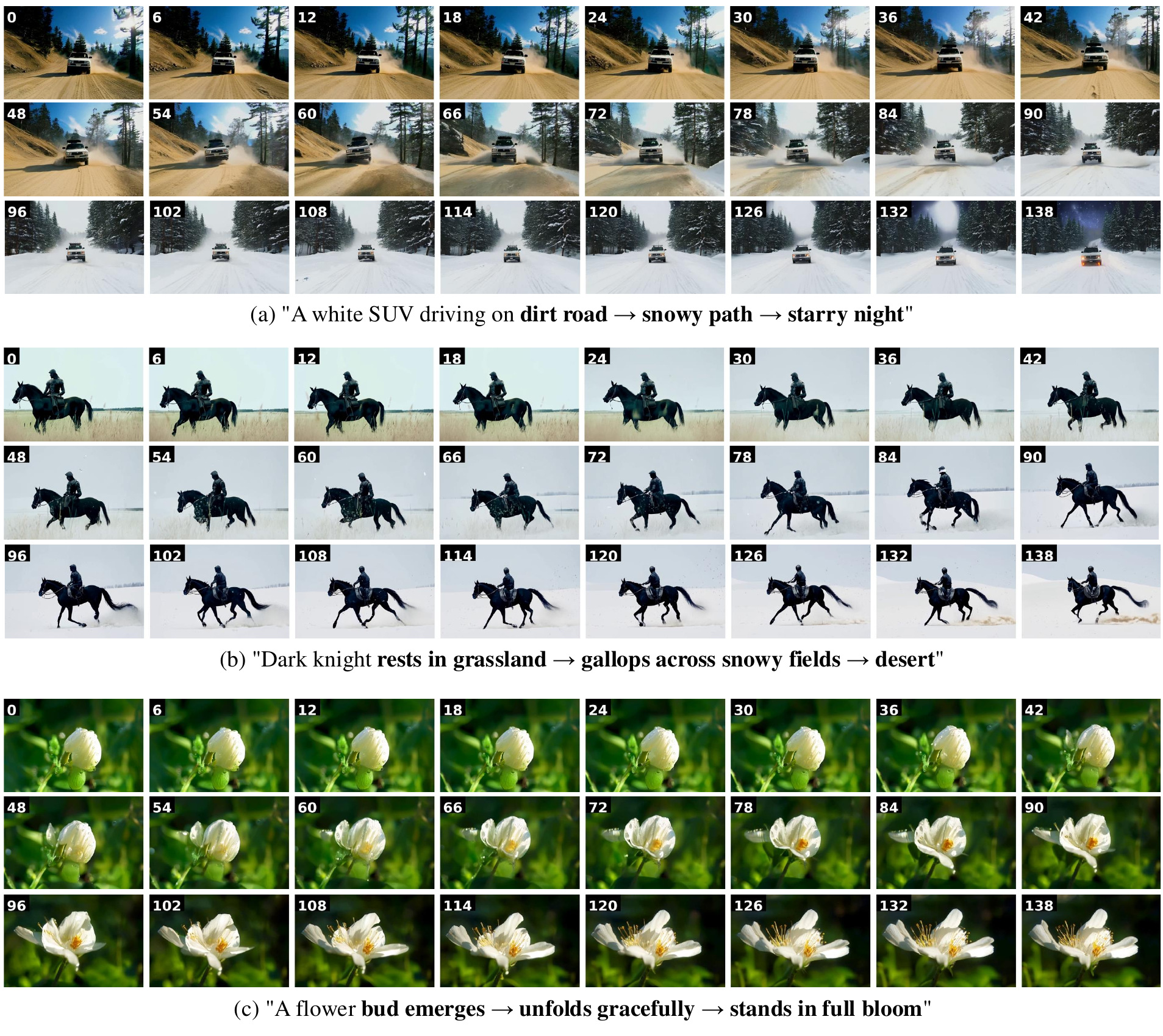}
    \vspace{-0.5em}
    \caption{More multi-prompt results} 
    \label{fig:more_result_2}
    \vspace{-1em}
\end{figure*}

\begin{figure*}
    \centering
    \includegraphics[width=0.95\linewidth]{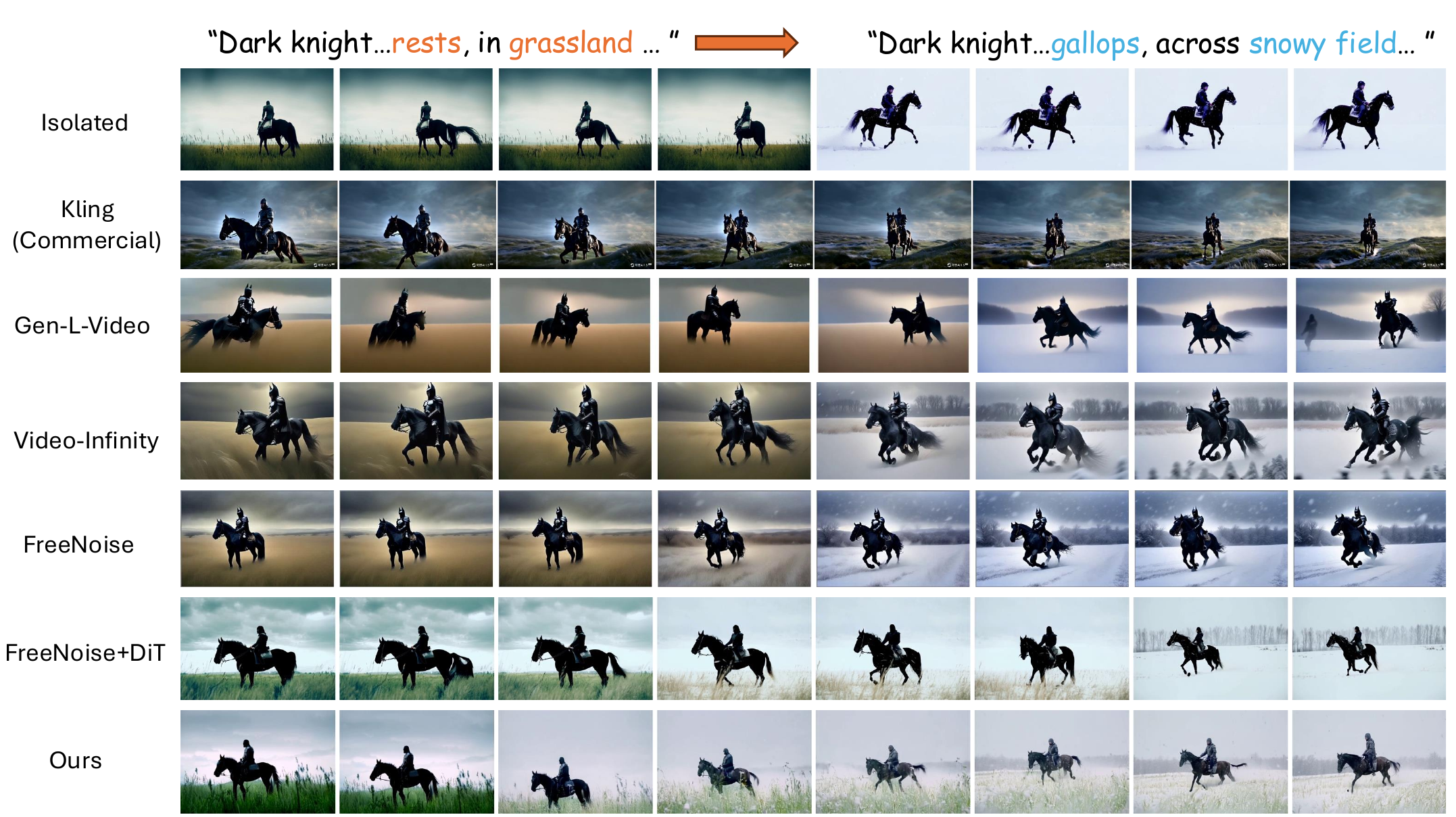}
    \vspace{-0.5em}
    \caption{Motion and background transition.} 
    \label{fig:fig_qualitative2}
    \vspace{-1em}
\end{figure*}

\begin{figure*}
    \centering
    \includegraphics[width=0.95\linewidth]{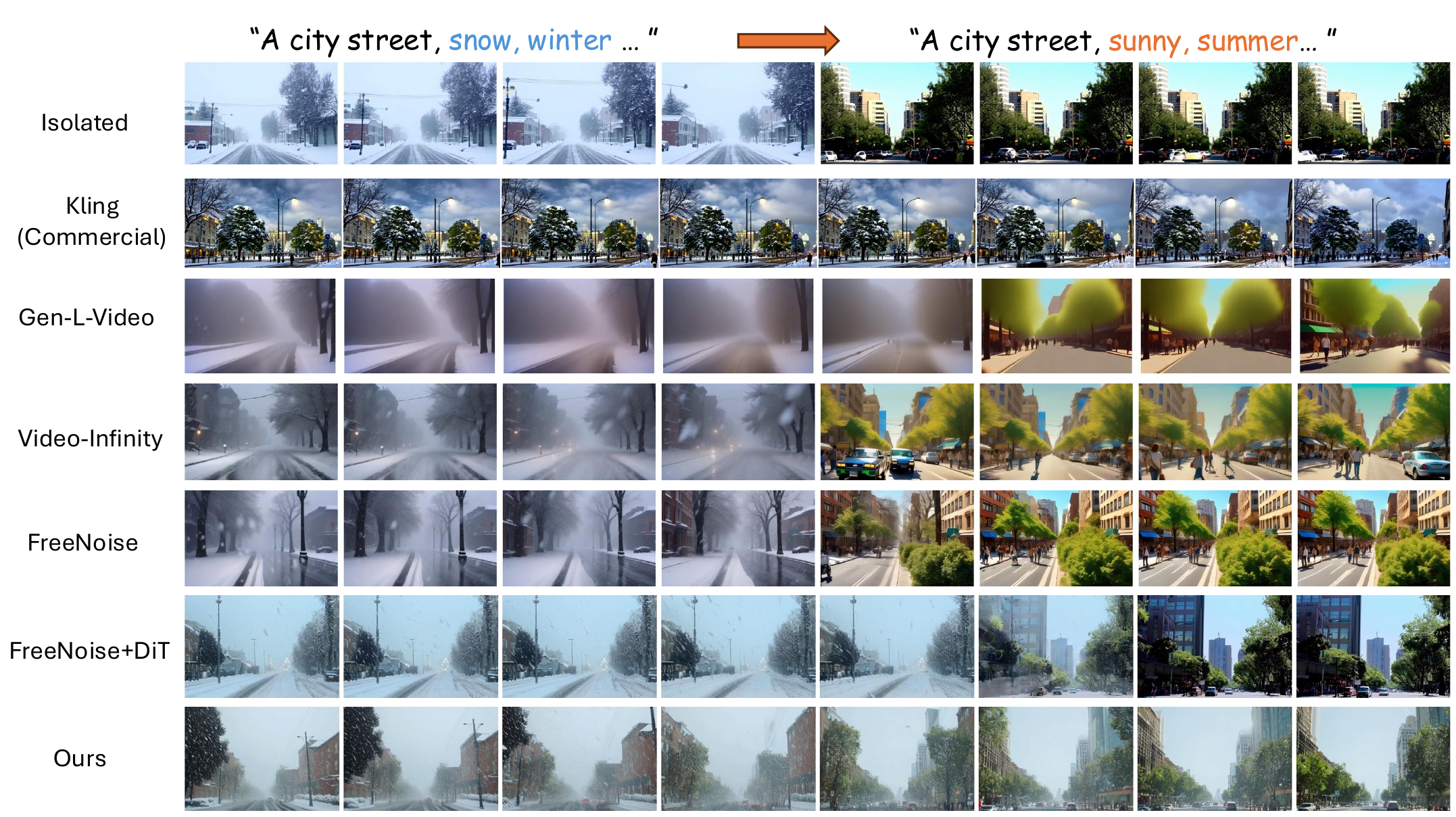}
    \vspace{-0.5em}
    \caption{Background transition.} 
    \label{fig:fig_qualitative3}
    \vspace{-1em}
\end{figure*}

\begin{figure*}
    \centering
    \includegraphics[width=0.95\linewidth]{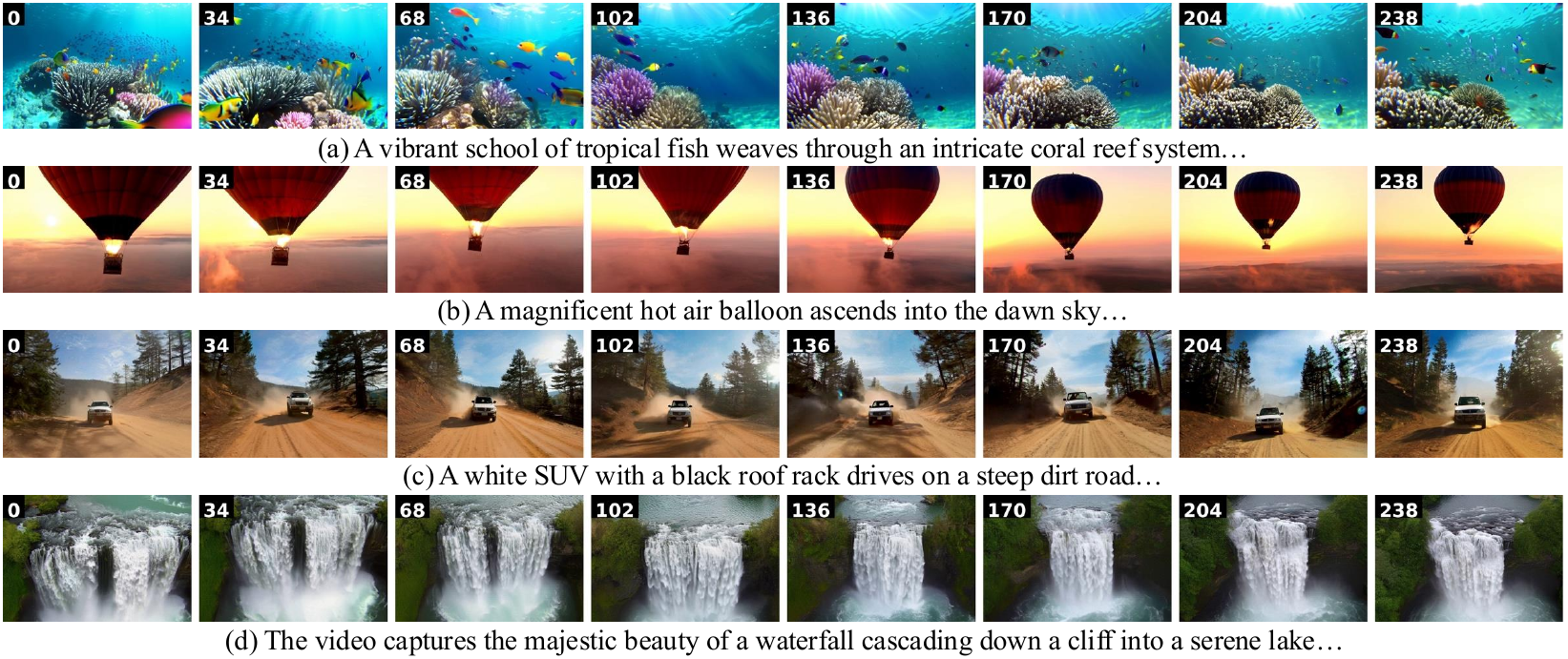}
    \vspace{-0.5em}
    \caption{Visualization of single prompt longer video generation.} 
    \label{fig:single_prompt_sup}
    \vspace{-1em}
\end{figure*}

\begin{figure*}
    \centering
    \includegraphics[width=0.95\linewidth]{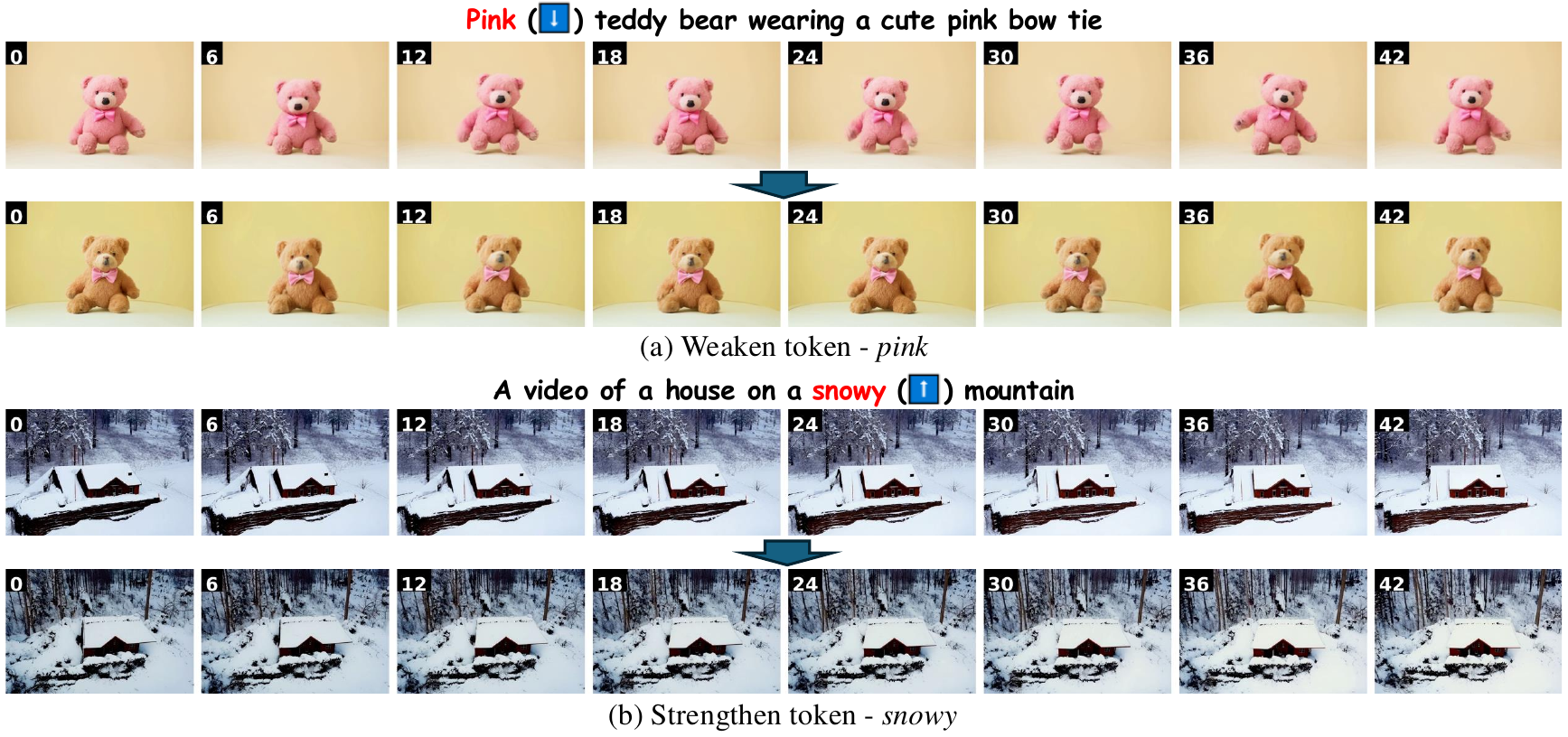}
    \vspace{-0.5em}
    \caption{Reweighting example of Video Editing.}
    \label{fig:reweight}
    \vspace{-1em}
\end{figure*}

\begin{figure*}
    \centering
    \includegraphics[width=0.95\linewidth]{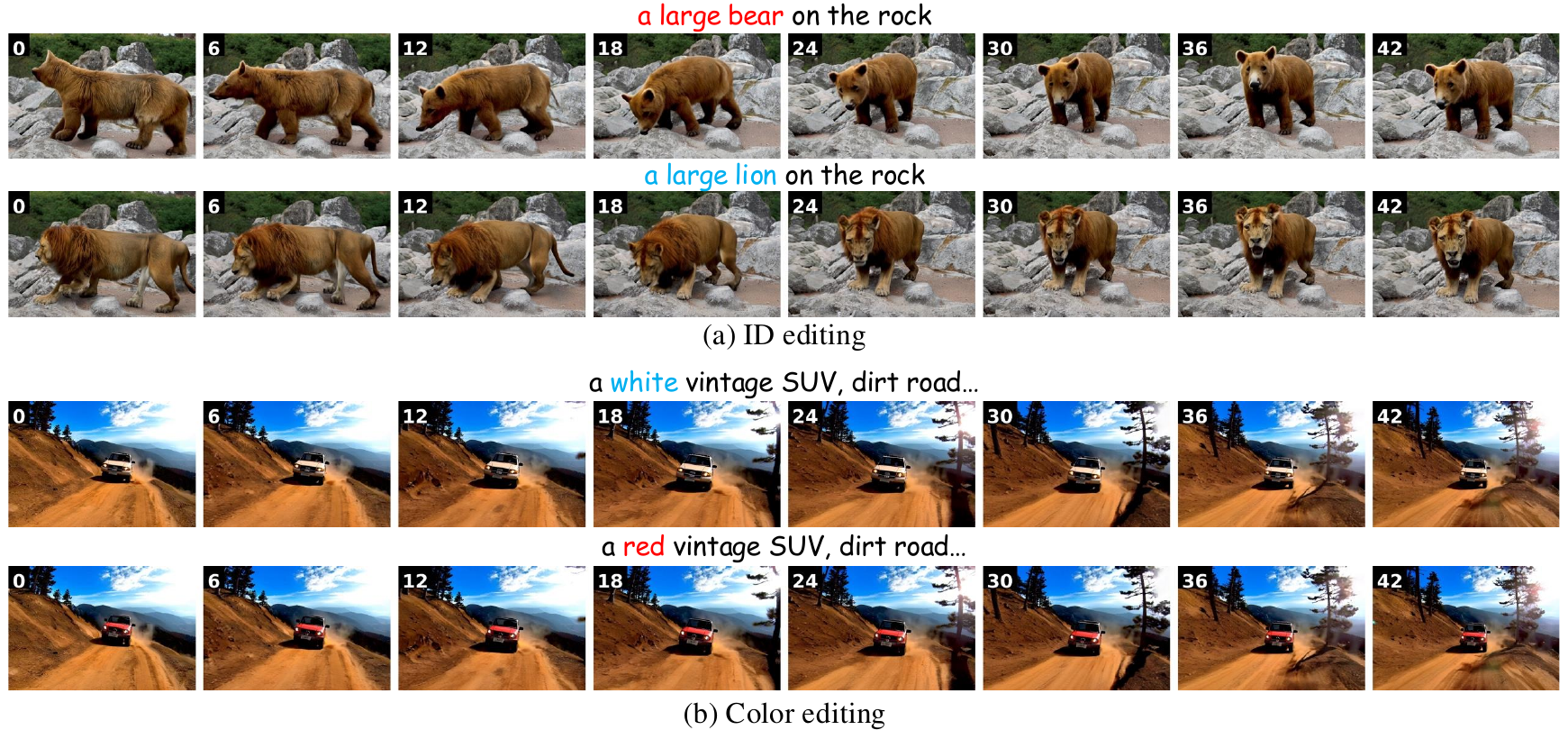}
    \vspace{-0.5em}
    \caption{Word Swap example of Video Editing.}
    \label{fig: word_swap}
    \vspace{-1em}
\end{figure*}

\begin{figure*}
    \centering
    \includegraphics[width=0.95\linewidth]{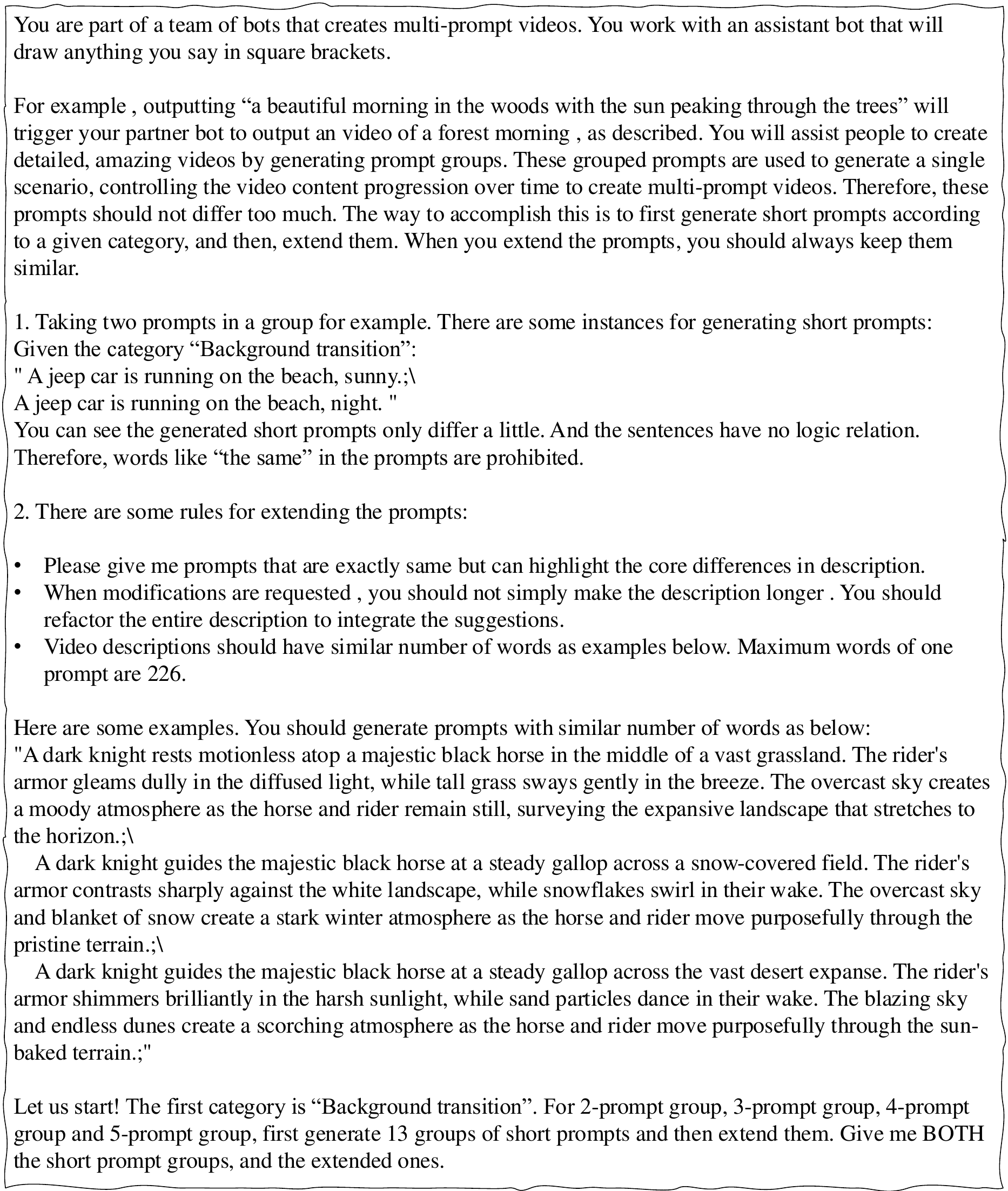}
    \vspace{-0.5em}
    \caption{Our instruction to create multiple individual long prompts based on short prompts group of specified types} 
    \label{fig:prompt_gen_instruction}
    \vspace{-1em}
\end{figure*} \fi

\end{document}